\documentclass[10pt,twocolumn,letterpaper]{article}

\usepackage{enumerate}
\usepackage{cvpr}
\usepackage{times}
\usepackage{diagbox}
\usepackage{pict2e}
\usepackage{colortbl}
\usepackage{float}
\usepackage{epsfig}
\usepackage{graphicx}
\usepackage{amsmath}
\usepackage{amssymb}
\usepackage{tabulary}
\usepackage{multirow}
\usepackage{subfigure}
\usepackage{soul}

\usepackage{epstopdf}
\usepackage{color}
\usepackage{caption}
\usepackage{verbatim}
\usepackage{array}
\usepackage{mathtools}
\usepackage{enumitem}
\usepackage{ragged2e}
\usepackage{booktabs}
\usepackage[british,UKenglish,USenglish,english,american]{babel}
\usepackage{amssymb}
\usepackage{amsthm}
\usepackage[table, dvipsnames]{xcolor}
\usepackage{subfigure}
\usepackage{amsmath}
\usepackage{ulem}

\usepackage[ruled]{algorithm2e}
\usepackage{algorithm,algpseudocode,float}
\usepackage{lipsum}
\definecolor{red}{rgb}{0.8,0.2,0.2}
\definecolor{blue}{rgb}{0.2,0.2,0.8}
\definecolor{yellow}{rgb}{1.0, 0.75, 0}
\definecolor{orange}{rgb}{1,0.5,0}


\usepackage{enumerate}
\usepackage{appendix}

\cvprfinalcopy 

\newlist{steps}{enumerate}{1}
\setlist[steps, 1]{label = Step \arabic*:}

\begin{document}

\title{Novel and Effective CNN-Based Binarization for Historically Degraded As-built Drawing Maps}

\author{Kuo-Liang Chung \qquad De-Wei Hsieh \\
\large National Taiwan University of Science and Technology \vspace{-.2em}\\
\large Department of Computer Science \& Information Engineering \vspace{-.2em}\\
\normalsize
\{klchung01,~hi201209910\}@gmail.com
}

\maketitle

\begin{abstract}
\vspace{-.5em}
Binarizing historically degraded as-built drawing (HDAD) maps is a new challenging job, especially in terms of removing the three artifacts, namely noise, the yellowing areas, and the folded lines, while preserving the foreground components well. In this paper, we first propose a semi-automatic labeling method to create the HDAD-pair dataset of which each HDAD-pair consists of one HDAD map and its binarized HDAD map. Based on the created training HDAD-pair dataset, we propose a convolutional neural network-based (CNN-based) binarization method to produce high-quality binarized HDAD maps. Based on the testing HDAD maps, the thorough experimental data demonstrated that in terms of the accuracy, PSNR (peak-signal-to-noise-ratio), and the perceptual effect of the binarized HDAD maps, our method substantially outperforms the nine existing binarization methods. In addition, with similar accuracy, the experimental results demonstrated the significant execution-time reduction merit of our method relative to the retrained version of the state-of-the-art CNN-based binarization methods.
\end{abstract}




\vspace{-1em}
\section{INTRODUCTION}
Historically degraded as-built drawing (HDAD) maps have been widely used to record the architectural and gauged foundation information. Among the considerable amount of HDAD maps, binarizing these HDAD maps is a necessary job prior to manipulating the binarized HDAD maps, and these image manipulations include optical character recognition (OCR), content retrieval, editing, data transmitting, compression, etc.

\begin{figure*}[]
  \centering
  \includegraphics[height=8cm]{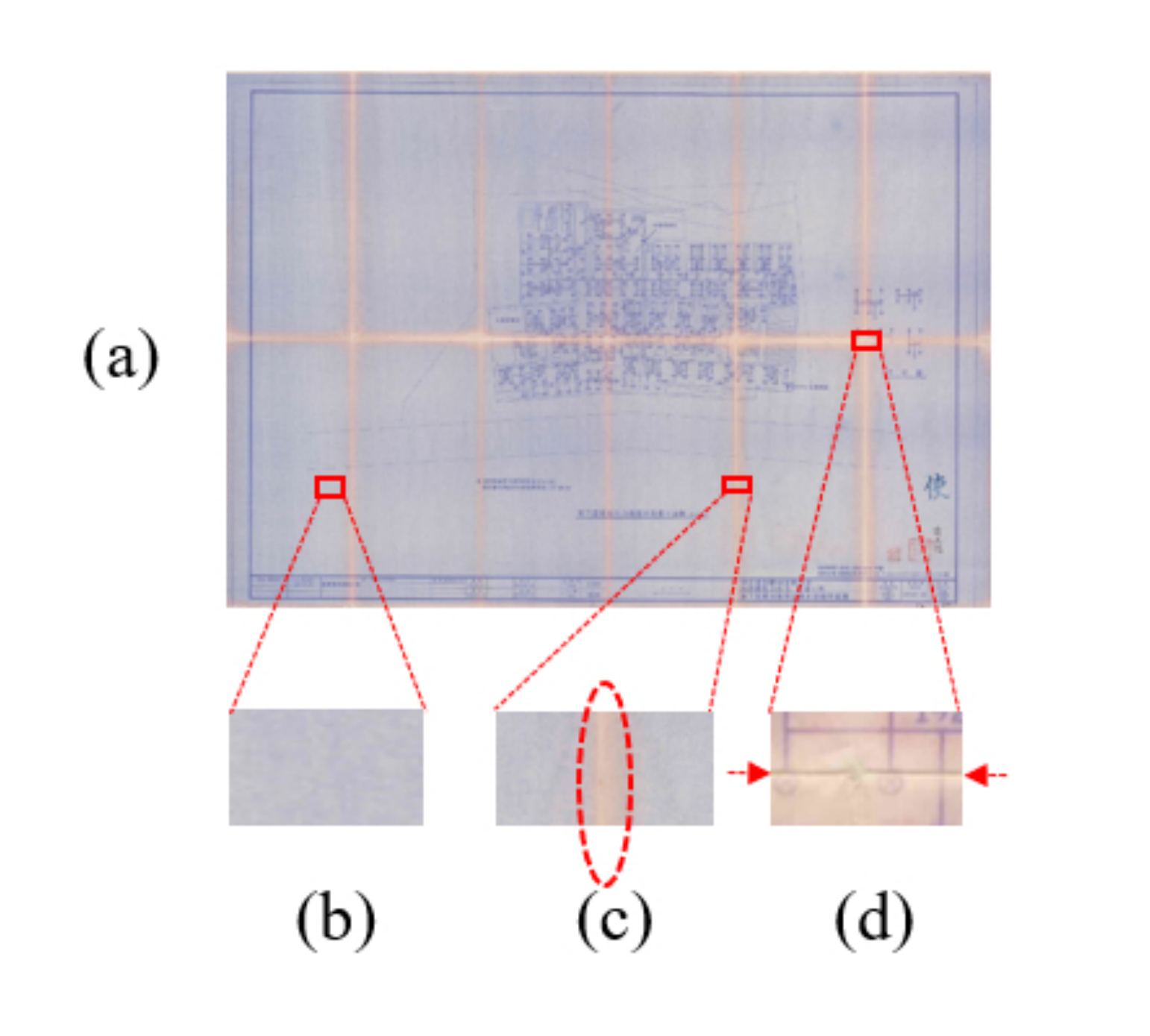}
  \caption{Three artifacts occurring in one HDAD map example. (a) The HDAD map. (b) Noise. (c) Yellowing area. (d) Folded line.}
  \label{fig:logo}
\end{figure*}

Generally, three kinds of artifacts always occur in HDAD maps, namely noise, the yellowing area, and the folded lines. One HDAD map example, as shown in Fig. \ref{fig:logo}(a), is taken to illustrate the three artifacts. As shown in Fig. \ref{fig:logo}(b), noise is often heavily scattered on the HDAD map. As highlighted by a red ellipse in Fig. \ref{fig:logo}(c), the yellowing area is caused due to moisture, oxidation, and aging. When removing the yellowing area, the foreground components covered by the yellowing area tend to be degraded. As highlighted by two arrows in Fig. \ref{fig:logo}(d), the folded line artifact is caused by folding the HDAD map, it is not easy to remove the folded line artifact well because the folded line is often misidentified as a line foreground. The above-mentioned three artifacts indicate that binarizing HDAD maps is a new challenging job.

In the past, many binarization methods were developed for degraded document maps and handwritten maps in which the two main artifacts considered are noise and the yellowing area. Although most of these methods work well for document maps and handwritten maps, but not for HDAD maps. Note that to alleviate the yellowing area influence before binarizing these maps, some color-to-gray transformations \cite{R. Hedjam}, \cite{A. S. Abutaleb}, \cite{M. Grundland}, \cite{A. A. Gooch}, \cite{K. Smith} were proposed as a preprocessing step. In the next subsection, the related binarization methods for document maps and handwritten maps are introduced.

\subsection{Related Work}\label{sec:IA}
  The related work on binarizing document maps and handwritten maps can be partitioned into three categories: the global thresholding-based binarization, the local thresholding-based binarization, and the deep learning-based binarization.\par

  \subsubsection{Global thresholding-based binarization}
  \quad The basic idea in this category is to determine a suitable threshold value as the pilot value to separate the foreground and the background in the map. Otsu \cite{N. Otsu} determined the threshold value such that the sum of two within-variances in the foreground and background is minimal, also achieving the maximal between-variance. Xue and Titterington \cite{J. H. Xue} showed the equivalence between Otsu’s method and the search for determining an optimal threshold value such that the absolute Student’s statistic is the largest.

  With a similar binarization performance as in Otsu’s method, Kapur $et$ $al$. \cite{J. N. Kapur} determined the threshold value such that the sum of the entropies in the foreground and background is maximal. Kavallieratou \cite{E. Kavallieratou} proposed an iterative histogram equalization-based global binarization (IHEGT) method. The experimental data demonstrated the binarization superiority of IHEGT over Otsu’s method and Sauvola and Pietikäinen's method \cite{J. Sauvola}.

  The global thresholding-based approach may suffer from the non-bimodel gray-value distribution of the considered image, making the foreground boundary disconnected. Due to the available codes, the two binarization methods in \cite{N. Otsu} and \cite{E. Kavallieratou} are included in the comparative methods.

  \subsubsection{Local thresholding-based binarization}
  \quad The local thresholding-based binarization approach determines the local threshold value of each pixel by considering the $w$$\times$$w$ window $W$ centered at that pixel. Niblack \cite{W. Niblack} determined the local threshold value, $T_{Niblack}(x,y)$, at location $(x, y)$ by the following formula:
  \begin{equation}
    \begin{aligned}
      T_{Niblack}(x,y)=m(x,y)+kS(x,y)
      \label{eq:differentiate1}
    \end{aligned}
  \end{equation}
  where ${m}$ and $S$ denote the mean gray value and the standard deviation of all pixels covered by $W$, respectively. ${k}$ is a user-specified parameter and is set to $0.1$ empirically. To improve Niblack's method, Sauvola and Pietikäinen \cite{J. Sauvola} determined the local threshold value by\\
  \begin{equation}
    \begin{aligned}
      T_{Sauvola}(x,y)=m(x,y)[1+k(S(x,y)/R-1)] \\
      \label{eq:differentiate2}
    \end{aligned}
  \end{equation}
  empirically, the values of $k$ and $R$ are set to $0.5$ and $128$, respectively.

  Howe \cite{N. R. Howe} first delivered a Markov random field based approach to label the initial foreground. Second, a data-fidelity energy function is formulated. Finally, the edge discontinuity idea is employed in the smoothness term of the energy function, and the experimental data indicated better quality relative to Gatos $et$ $al$.'s method \cite{B. Gatos}.

  Chiu $et$ $al$. \cite{Y. H. Chiu} first applied a statistical approach to determine the two thresholds, $T_f$ and $T_b$, such that the threshold $T_f$ leads to the largest relative increasing rate of newly generated foreground pixels. Jia $et$ $al$. \cite{SSP} utilized an edge map to estimate the stroke-edges and applied the multiple-threshold voting method to decide the suitable local threshold value. Su $et$ $al$. \cite{B. Su} employed the contrast and edge information in their local thresholding-based binarization method for binarizing degraded document maps.

  The local thresholding-based approach may suffer from the hollow foreground interior. Due to the available codes, the five local thresholding-based binarization methods in \cite{W. Niblack}, \cite{J. Sauvola}, \cite{N. R. Howe}, \cite{Y. H. Chiu}, \cite{SSP} are included in the comparative methods.

  \subsubsection{Deep learning-based binarization}
   \quad Using the convolutional neural networks (CNN) \cite{Y. LeCun}, Vo $et$ $al$. \cite{Q. N. Vo} proposed a hierarchical deep supervised network (DSN)-based binarization method for handwritten maps. Calvo-Zaragoza $et$ $al$. \cite{J. Calvo-Zaragoza} proposed a pixel- and CNN-based method for binarizing the musical document maps. Furthermore, based on the DIBCO training dataset \cite{DIBCO} and the convolutional auto-encoder networks, Calvo-Zaragoza and Gallego \cite{J. Calvo} proposed an effective binarization method for document images; experimental data demonstrated that their method is superior to the previous methods in \cite{N. Otsu}, \cite{W. Niblack}, \cite{J. Sauvola}, \cite{R. C. Gonzalez}, \cite{E. Kavallieratou}, \cite{B. Su}, \cite{J. Pastor-Pellicer}.

   Using the fully convolutional networks (FCN) \cite{J. Long} with four hierarchical scales, Tensmeyer and Martinez \cite{C. Tensmeyer} proposed a FCN-based binarization method for document maps. Zhao $et$ $al$. \cite{J. Zhao} proposed a generative adversarial network (GAN)-based method to binarize the handwritten documents, and the experimental data indicated better binarization quality relative to some previous methods.

  Due to the available codes, the two CNN-based methods in \cite{J. Calvo}, \cite{J. Zhao} are included in the comparative methods.

  \subsection{Motivations} \label{sec:IB}
  Because the size of each HDAD map is huge and is ranged from 2436$\times$1738 to 10124$\times$6962, to reduce human efforts on labeling each pixel annotation as a background pixel or a foreground pixel, our first motivation is to design a fast and effective labeling method to create the HDAD-pair dataset, in which each HDAD-pair consists of one input HDAD map and its ground truth binarized HDAD map. Our second motivation is to design the first effective CNN-based binarization method for HDAD maps such that in terms of the accuracy, PSNR (peak-signal-to-noise-ratio), and the perceptual effect of the binarized HDAD maps, our method can outperform the above-mentioned nine binarization methods \cite{N. Otsu}, \cite{W. Niblack}, \cite{J. Sauvola}, \cite{E. Kavallieratou}, \cite{N. R. Howe}, \cite{Y. H. Chiu}, \cite{SSP}, \cite{J. Calvo}, \cite{J. Zhao}. Here, the perceptual effect of the binarized HDAD maps are subjectively evaluated by human eyes. Based on our created HDAD-pair dataset, our third motivation is to retrain the two state-of-the-art methods \cite{J. Calvo}, \cite{J. Zhao}, and then report the performance merit of our method.

  \subsection{Contributions} \label{sec:IC}
  To address our three motivations, in this paper, we first propose a semi-automatic labeling method to create the HDAD-pair dataset. Then, based on the created HDAD dataset, we propose an effective CNN-based binarization method for HDAD maps. The three contributions of this paper are described as follows.

  In the first contribution, we propose a semi-automatic labeling method to create the HDAD-pair dataset. For $I^{HDAD}$, we first propose a fusion approach to integrate our modified local-thresholding (MLT) based binarization method, which will be presented in Subsection \ref{sec:II1}, and the IHEGT method \cite{E. Kavallieratou}, which will be revisited in Subsection \ref{sec:IIIA2}, to construct a rough binarized HDAD map, namely $I^{bin,HDAD}_{rough}$. Here, the rough binarized HDAD map inherits the hole-free merit in the foreground interior by using IHEGT and the connected boundary of foreground merit by using our MLT method. Next, to preserve the foreground lines and remove noise, a center weighted median filter based noise removal process is applied to delete the sparse noise in $I^{bin,HDAD}_{rough}$. Furthermore, a slight handmade adjustment is applied to produce the ground truth binarized HDAD map, creating the HDAD-pair dataset.

  In the second contribution, based on the created HDAD-pair dataset, we propose a novel and effective CNN-based binarization method for HDAD maps, and the proposed method achieves substantial accuracy, PSNR, and the perceptual effect improvements relative to the nine comparative methods.

  In the third contribution, we randomly select 62 HDAD-pairs from the newly created dataset \cite{website}, as the training set, and we randomly select 12 exclusive HDAD-pairs from the dataset as the testing set. The comprehensive experimental data demonstrated that in terms of four accuracy metrics, namely recall, specificity, precision, and F-measure, PSNR, and the perceptual effect, our binarization method is clearly superior to the considered nine binarization methods \cite{N. Otsu}, \cite{W. Niblack}, \cite{J. Sauvola}, \cite{N. R. Howe}, \cite{E. Kavallieratou}, \cite{Y. H. Chiu}, \cite{SSP}, \cite{J. Calvo}, \cite{J. Zhao}. In addition, with the similar accuracy and PSNR, the experimental data demonstrated that in terms of execution time and the number of parameters used in the considered CNN frameworks, our method clearly outperforms the retrained version of the two state-of-the-art methods \cite{J. Calvo}, \cite{J. Zhao}.

  The rest of this paper is organized as follows. In Section II, we propose a semi-automatic labeling method to create the new HDAD-pair dataset effectively. In Section III, based on the created dataset, we propose an effective CNN-based binarization method for HDAD maps. In Section IV, the comprehensive experimental results are reported to demonstrate the accuracy, PSNR, the perceptual effect, and the execution-time merits of our method. In Section VI, some concluding remarks are addressed.\\

\section{The proposed semi-automatic labeling method to generate the HDAD-pair dataset}

   \label{sec:II}
  In this section, a new semi-automatic labeling method is proposed to label each pixel of the HDAD map to be a foreground pixel or a background pixel, thus producing the HDAD-pair dataset. The proposed labeling method consists of two stages. In the first stage, a fusion-based approach is proposed to produce the rough HDAD-pair dataset. In the second stage, the center weighted median filter technique for removing noise and a slight handmade adjustment are applied to refine the binarized HDAD maps, creating the resultant HDAD-pair dataset. For convenience, the refined binarized HDAD map is denoted by $I^{bin,HDAD}_{fine}$. In our semi-automatic labeling method, the first stage and the center weighted median filter-based noise removal in the second stage are totally automatic.

  \subsection{The First Stage: The proposed fusion-based approach to generate the rough HDAD-pair Dataset} \label{sec:IIA}
  In this subsection, we first present our MLT method, and then revisit the IHEGT method \cite{E. Kavallieratou}. Furthermore, a fusion approach, which integrates MLT and IHEGT together, is proposed to generate the rough HDAD-pair dataset automatically.

  \subsubsection{The proposed MLT method} \label{sec:II1}
  \quad Differing from two thresholds used in Chiu $et$ $al$.'s method \cite{Y. H. Chiu}, our MLT method uses only one threshold to separate the foreground pixel and the background pixel. Our MLT method is based on the idea: if the HDAD gray pixel-value at location $(x,y)$, denoted by $H_g(x,y)$, is lower than the mean gray-value of the block $H^b$ covered by the $w$$\times$$w$ window $W$ centered at location $(x,y)$ and the gradient-value of $H_g(x,y)$ is higher than the mean gradient-value of the block $H^b$, $H_g(x,y)$ tends to be a foreground pixel; otherwise, $H_g(x,y)$ tends to be a background pixel. Therefore, the threshold value used in MLT is defined by

  \begin{equation}
    \label{eq:Chiu_method}
    \begin{aligned}
      T_{MLT (x; y)} = \mu_{H_g (x,y)}\left (\ 1-ke^{-\mu_{\bigtriangledown H_g (x, y)}/M} \right )
    \end{aligned}
  \end{equation}
  where $H_g$ denotes the 256$\times$256 gray HDAD block covered by a 256$\times$256 window centered at the location $(x, y)$. $\mu_{H_g\left ( x, y \right )}$ and $\mu_{\bigtriangledown H_g\left ( x, y \right )}$ denote the mean gray-value and the mean gradient-value of $H_g$, respectively; $M$ denotes the maximal gradient-value in $H_g$ and $M$ is used to normalize the mean gradient-value to be in the range $[0, 1]$. Empirically, the best choices of $k$ and $w$ are set to 0.02 and 17, respectively.

  \begin{figure*}
      \centering
            \subfigure[]{\includegraphics[scale = 0.25]{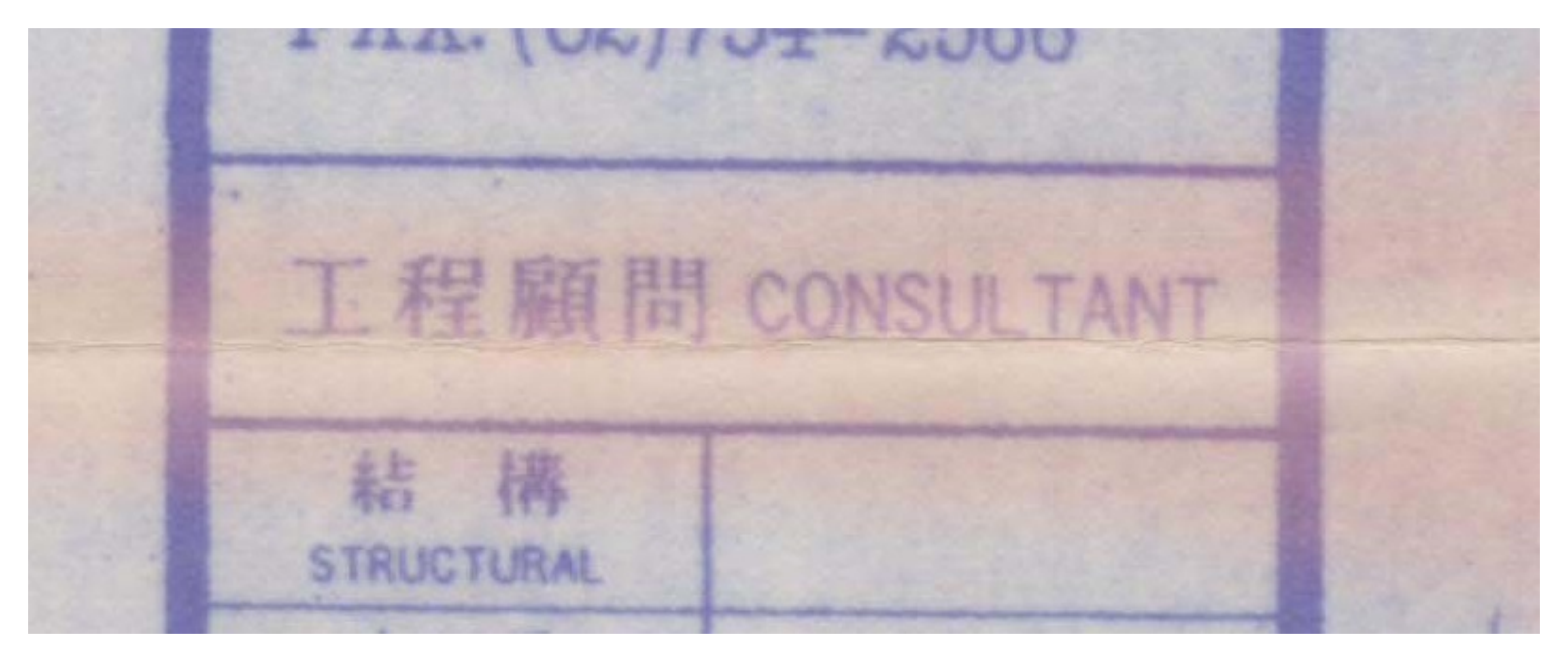}}
            \subfigure[]{\includegraphics[scale = 0.27]{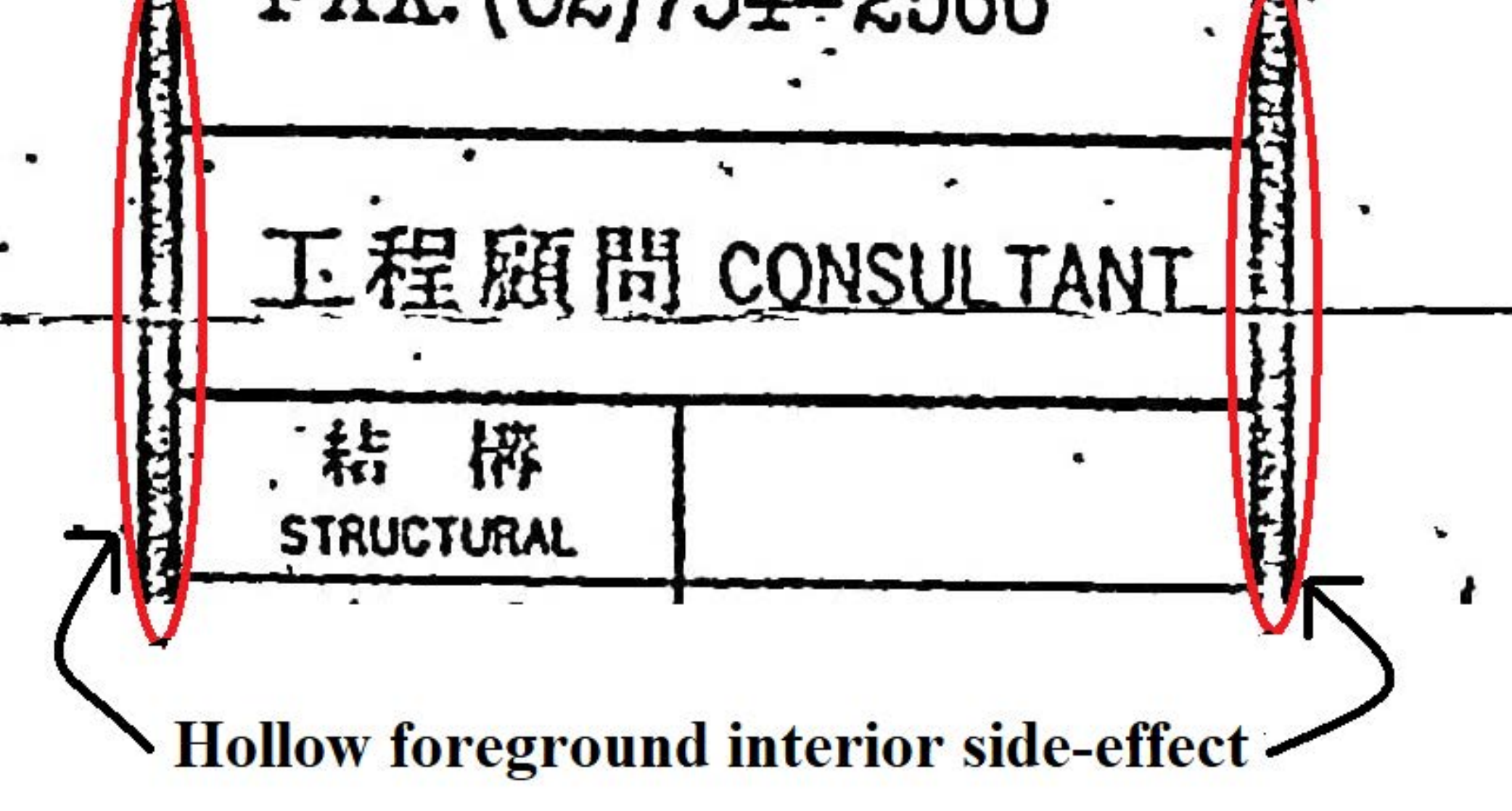}}
        \caption{Hollow foreground interior side-effect of the proposed MLT method. (a) The given HDAD map. (b) The binarized HDAD map by the proposed MLT method.}
        \label{fig:four_pic}
  \end{figure*}

  Since our MLT method does not include the region growing technique used in \cite{Y. H. Chiu}, it leads to low computation merit while preserving similar binarization performance relative to the method in \cite{Y. H. Chiu}. Although the proposed MLT method has good binarization performance, it sometimes suffers from the hollow foreground interior side-effect for fat foreground components, as shown in Fig. \ref{fig:four_pic}(b).

  \subsubsection{Revisiting IHEGT} \label{sec:IIIA2}
  \quad As mentioned before, IHEGT \cite{E. Kavallieratou} has hole-free advantage in the foreground interior but sometimes has the disconnected foreground boundary side-effect as shown in Fig. \ref{fig:refine_handmade}(a). In our fusion approach, which will be described in the next subsection, the binarized results by using our MLT method and by using IHEGT are integrated to combine the connected foreground boundary merit of MLT and the hole-free foreground interior advantage of IHEGT.

  We revisit IHEGT \cite{E. Kavallieratou} and rewrite it as the following four-step procedure:

  \begin{steps}[leftmargin=*]
    \item The average gray value of the gray HDAD map $H_{g}$, $\mu_{H_{g}}$, is calculated.
    \item Perform the assignment operation $H_g{(x,y)}$ = 255 + $H_g{(x,y)}$ - $\mu_{H_{g}}$. If $H_g{(x,y)}$ $\geq 255$, we set $H_g{(x,y)}$ to 255, denoting the background pixel; otherwise, we do nothing.
    \item To perform the histogram equalization on $H_g$, the value of $H_g{(x,y)}$ is mapped to
    \begin{align}
      \begin{split}
      \label{eq:Global_method}
        H_{g}{(x, y)}=255 -255 *(\frac{255 - H_g\left ( x, y \right )}{255 - min_{H_{g}}})
      \end{split}
    \end{align}

    where `$min_{H_{g}}$' denotes the minimal pixel-value in $H_{g}$.
    \item If two average gray values in consecutive iterations are equal, it means that all the background pixels with value 255 have been filtered out and the current map contains only foreground pixels; we thus stop IHEGT; otherwise, we go to Step 1.
  \end{steps}

   \subsubsection{The proposed fusion-based approach to produce the rough HDAD-pair dataset}
   \label{sec:IIIC}

   \quad In this subsection, the proposed fusion-based approach is used to integrate MLT and IHEGT together for producing a rough HDAD-pair dataset.

   Let the binarized HDAD map by using our MLT method be denoted by $I^{bin, HDAD}_{MLT}$ and let that by using IHEGT be denoted by $I^{bin, HDAD}_{IHEGT}$. Using the proposed fusion approach, the rough binarized HDAD map is expressed by

   \begin{equation}
     I^{bin, HDAD}_{rough} = I^{bin, HDAD}_{MLT} \cup I^{bin, HDAD}_{IHEGT}
     \label{eq:pr}
   \end{equation}
  Using the same HDAD map example in Fig. \ref{fig:four_pic}(a), $I^{bin,HDAD}_{MLT}$ and $I^{bin,HDAD}_{IHEGT}$ have been depicted in Fig. \ref{fig:four_pic}(b) and Fig. \ref{fig:refine_handmade}(a), respectively. By Eq. (\ref{eq:pr}), the rough binarized HDAD map is depicted in Fig. \ref{fig:refine_handmade}(b). From Fig. \ref{fig:refine_handmade}(b), although the side-effects of MLT and IHEGT have been alleviated by using our fusion approach, there are a few scattered noise and imperfect foreground components caused by removing folded lines in the HDAD map. Fortunately, the side-effects of MLT and IHEGT have been alleviated.

  \subsection{The second stage: refinement process to generate the fine HDAD-Pair dataset} 
\begin{figure}
      \centering
          \subfigure[]{\includegraphics[scale = 0.27]{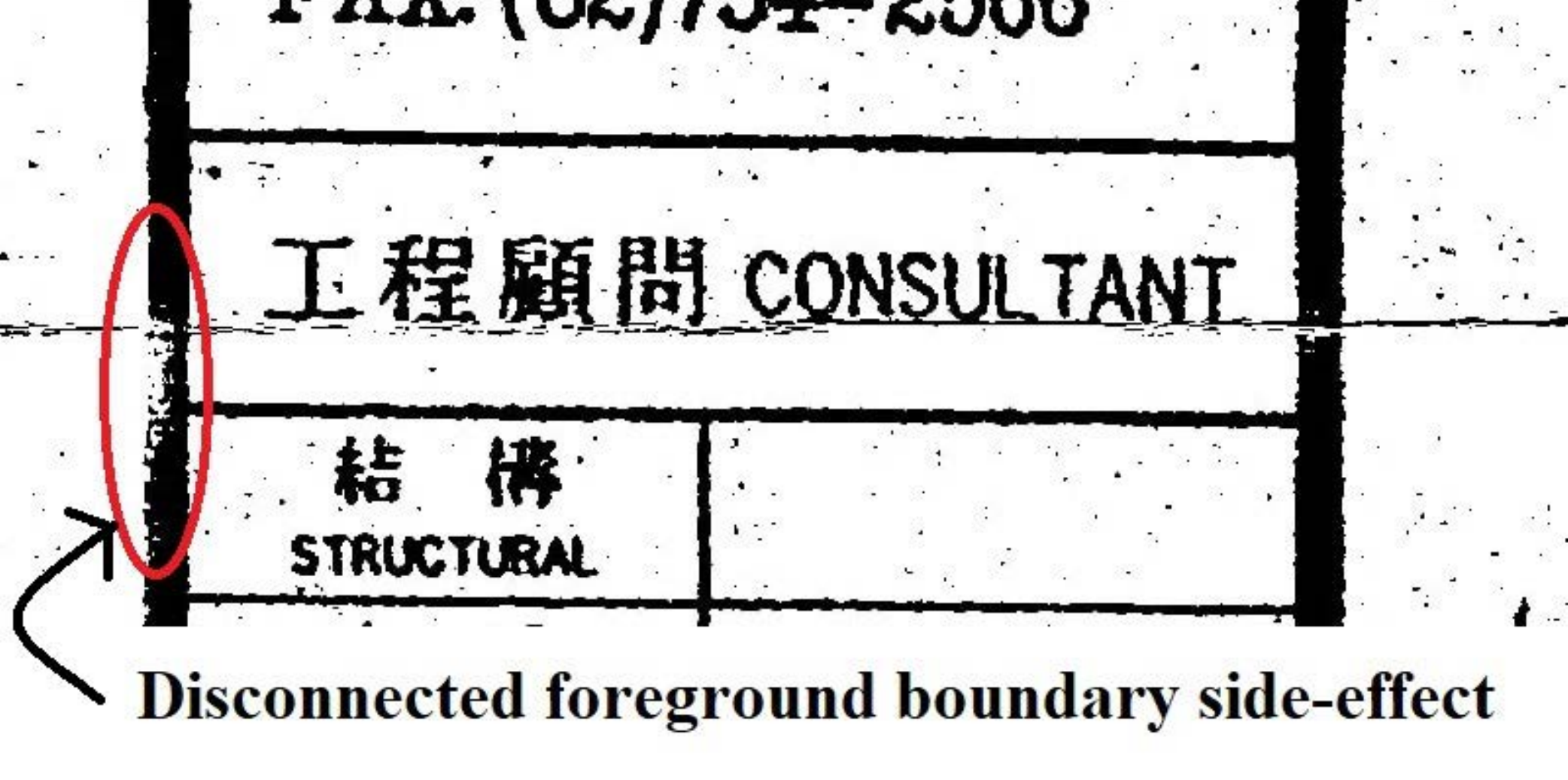}}
      \centering
          \subfigure[]{\includegraphics[scale = 0.27]{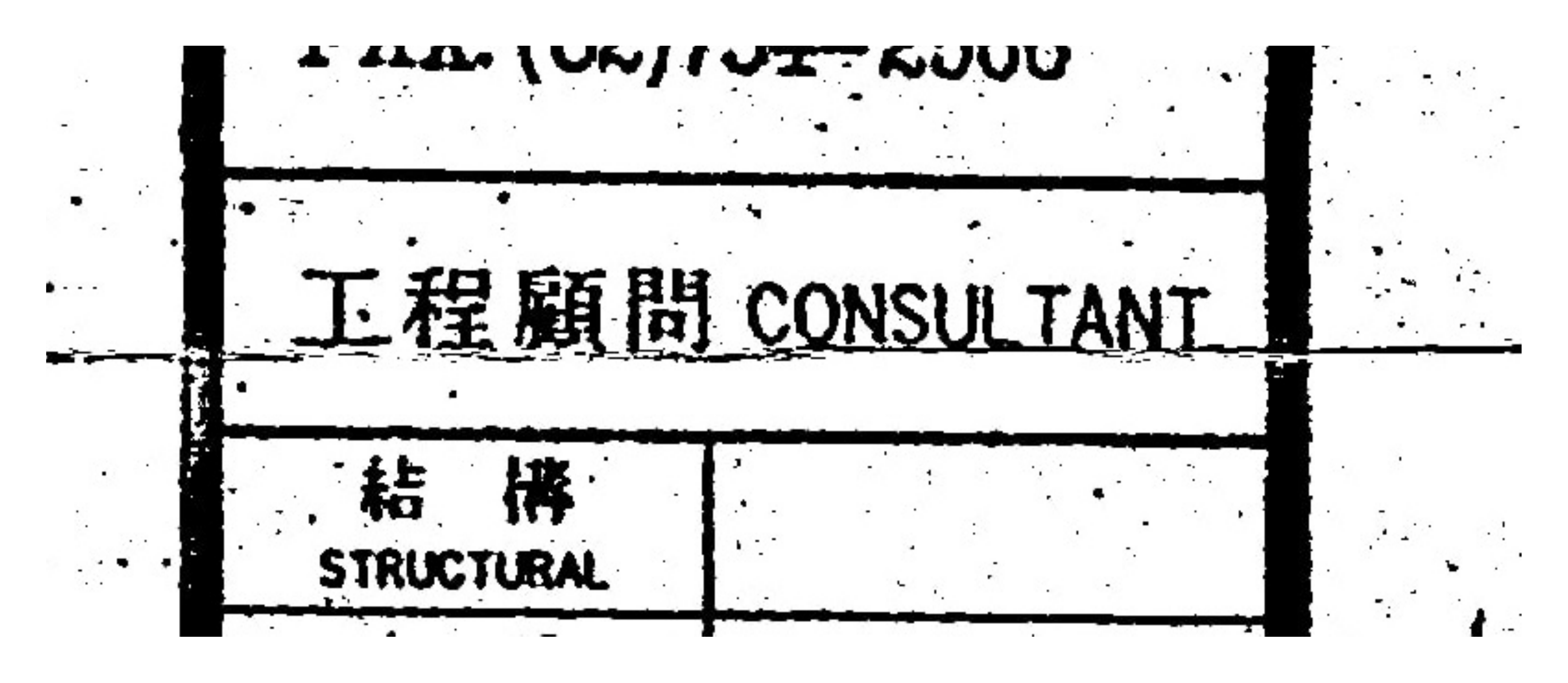}}
      \centering
          \subfigure[]{\includegraphics[scale = 0.205]{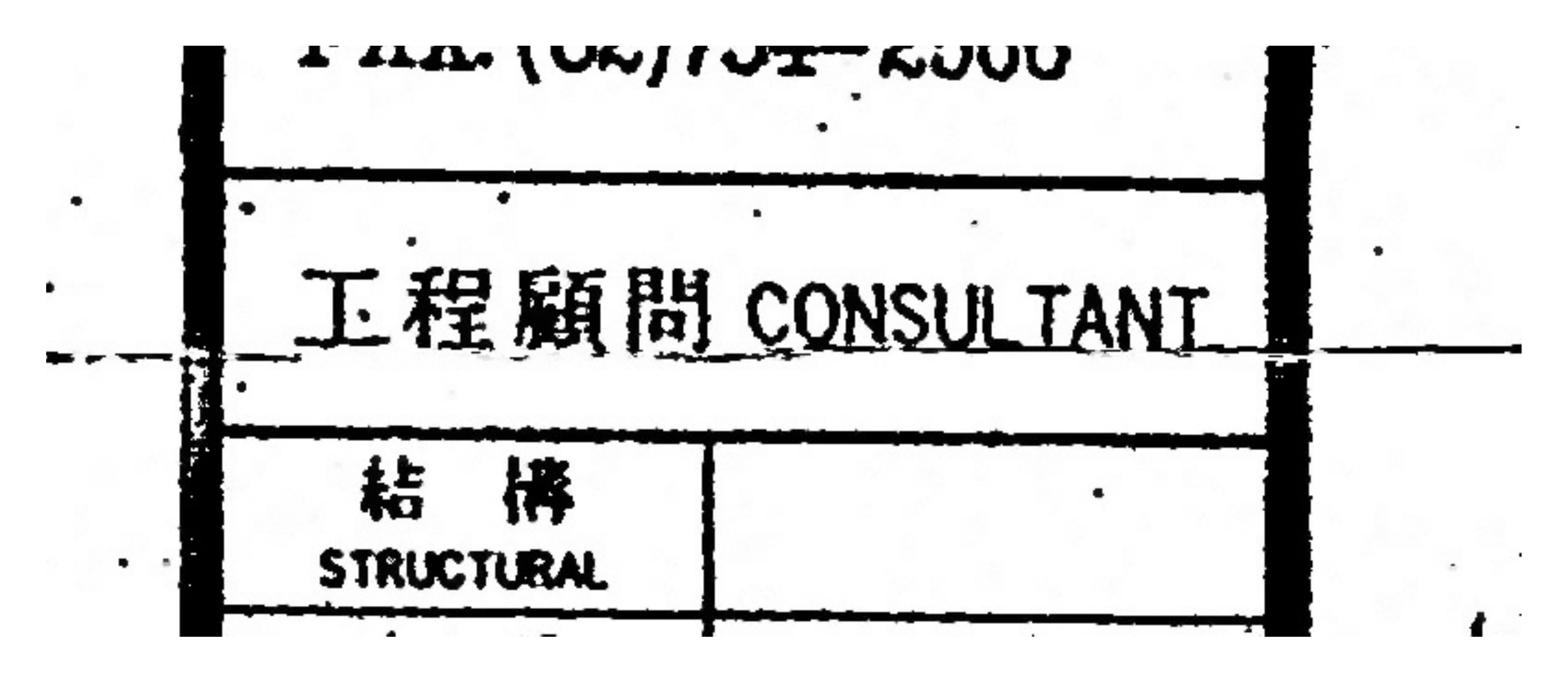}}
      \centering
          \subfigure[]{\includegraphics[scale = 0.27]{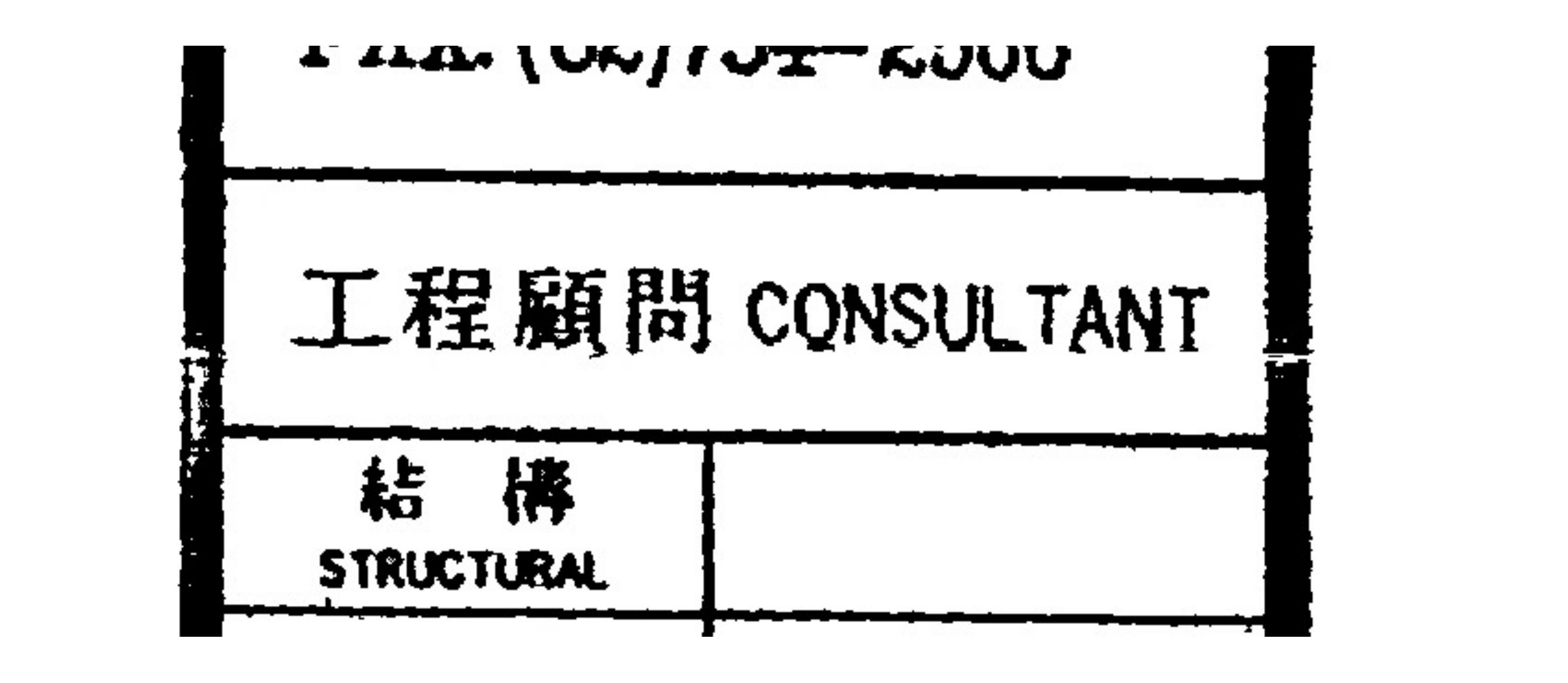}}
      \caption{(a) Disconnected foreground boundary side-effect of the IHEGT method for Fig. \ref{fig:four_pic}(a). (b) The rough binarized HDAD map by performing the proposed fusion-based approach on Fig. \ref{fig:four_pic}(a). (c) The refined binarized HDAD map by performing the center weighted median filter-based noise removal process on Fig. \ref{fig:refine_handmade}(b). (d) The finally binarized HDAD map by a slight handmade adjustment on Fig. \ref{fig:refine_handmade}(c).}
      \label{fig:refine_handmade}
  \end{figure}

  To remove the remaining scattered noise in the rough binarized HDAD map and to avoid disconnecting the thin foreground line, the center weighted median filter-based noise removal process with a 7$\times$7 mask \cite{S. J. Ko} is performed on $I^{bin, HDAD}_{rough}$, as shown in Fig. \ref{fig:refine_handmade}(b), and Fig. \ref{fig:refine_handmade}(c) depicts the refined binarized HDAD map. After a slight handmade adjustment to refine Fig. \ref{fig:refine_handmade}(c), Fig. \ref{fig:refine_handmade}(d) illustrates the finally binarized HDAD map which will be included in the fine HDAD-pair dataset.

\section{The proposed CNN-based binarization method for HDAD maps} \label{sec:III} 
In this section, we first describe the CNN framework used in our binarization method. Next, based on the newly created HDAD-pair dataset, the training step is described and the three novelties in our CNN framework are highlighted to explain why the number of parameters used is much less than that in the two state-of-the-art methods in \cite{J. Calvo}, \cite{J. Zhao}, leading to the execution-time reduction merit of our method. Finally, the testing step is presented.

\begin{table}[]  
\centering
\caption{The detailed configuration of CNN used in our binarization method.}
\begin{tabular*}{7.5cm}{c c c c  }%
\toprule
Layer   & Kernel Size & Stride & Feature Map   \\ \hline
\midrule
CONV1\_1 & 3x3x32     & 1      & 224x224x32     \\ \hline
CONV1\_2 & 3x3x32     & 2      & 112x112x32     \\ \hline
CONV2\_1 & 3x3x32     & 1      & 112x112x32     \\ \hline
CONV2\_2 & 3x3x32     & 2      & 56x56x32       \\ \hline
CONV3\_1 & 3x3x32     & 1      & 56x56x32       \\ \hline
CONV3\_2 & 3x3x32     & 2      & 28x28x32       \\ \hline
CONV4\_1 & 3x3x32     & 1      & 28x28x32       \\ \hline
CONV4\_2 & 3x3x32     & 2      & 14x14x32       \\ \hline
CONV5\_1 & 3x3x32     & 1      & 14x14x32       \\ \hline
CONV5\_2 & 3x3x32     & 2      & 7x7x32         \\ \hline
DECONV1 & 3x3x2     & 2      & 14x14x2        \\ \hline
DECONV2 & 3x3x2     & 2      & 28x28x2        \\ \hline
DECONV3 & 3x3x2     & 2      & 56x56x2        \\ \hline
DECONV4 & 3x3x2     & 2      & 112x112x2      \\ \hline
DECONV5 & 3x3x2     & 2      & 224x224x2      \\ \hline
\bottomrule
\label{table:EEFCN_config}
\end{tabular*}
\end{table}

\begin{figure*}[!h]
  \centering
  \includegraphics[scale=0.26]{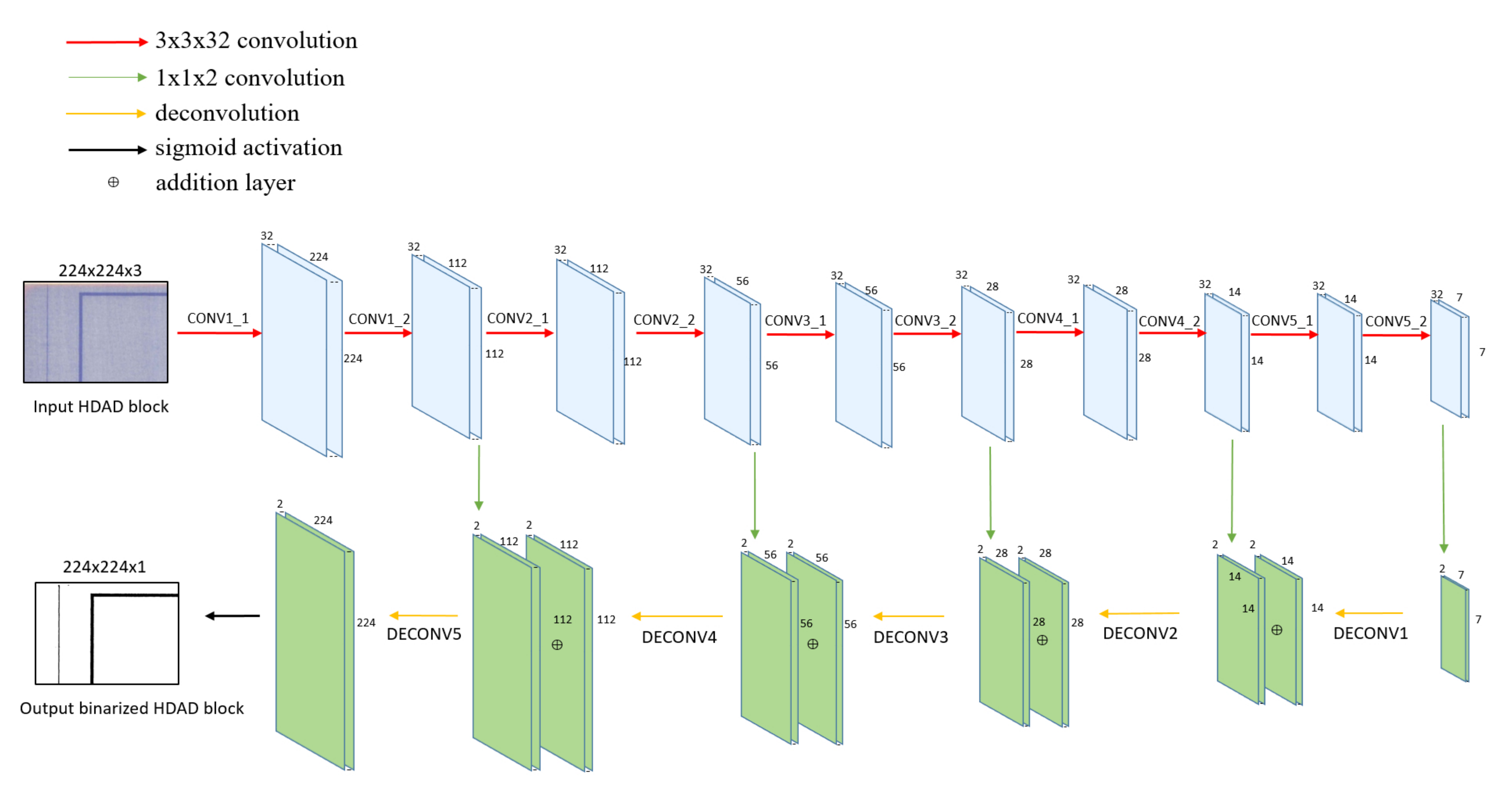}
  \caption{The proposed CNN framework for binarizing HDAD maps.}
  \label{fig:deep_learning_arc}
\end{figure*}

The CNN framework used in our binarization method is depicted in Fig. \ref{fig:deep_learning_arc}, and its configuration is illustrated in Table \ref{table:EEFCN_config}. In Fig. \ref{fig:deep_learning_arc}, the ten convolutional layers are denoted by CONV1\_1, CONV1\_2, ..., and CONV5\_2 in the upper part of Fig. \ref{fig:deep_learning_arc}, and the five deconvolutional layers, which are denoted by DECONV1, DECONV2, ..., and DECONV5 in the lower part of Fig. \ref{fig:deep_learning_arc}.

There are three novelties in the design of our CNN for binarization. In the first novelty, instead of the pooling layer used in \cite{O. Ronneberger}, each convolutional layer with even index in Fig. \ref{fig:deep_learning_arc}, we set its stride to be 2, leading to the reduction of the number of parameters which includes the total number of kernel weights and total number of bias-weights required in all convolutional layers. In the second novelty, because of only two classes, i.e. foreground and background, considered in our study, to reduce the $7\times7\times32$ feature map (see the rightmost feature map in the upper part of Fig. \ref{fig:deep_learning_arc}) to a $7\times7\times2$ feature map, we perform a $1\times1\times2$ convolution operation on the $7\times7\times32$ feature map. In the same way, all the feature maps associated with even indices in the upper part of Fig. \ref{fig:deep_learning_arc} are reduced to much smaller feature maps, each with only two pieces. In the third novelty, the above feature map reduction effect leads to decreasing the number of total parameters. In Fig. \ref{fig:deep_learning_arc}, each deconvolution operation upsamples the input feature map to the four times feature map.

In the training step, we adopt the binary cross-entropy as our loss function, and ``Adam'' as our optimizer. each training HDAD map is first partitioned into a set of $224\times224$ HDAD blocks. Then, each HDAD block and the corresponding ground truth binarized HDAD block form an end-to-end training block-pair. After processing all the 62 training HDAD-pair maps and performing 50 epochs, the loss value is less than 0.03.

~\\
\indent In the testing step, each testing HDAD map, which is disjointed from that in the training dataset, is fed into our CNN-based binarization method. First, our method automatically partitions the testing HDAD map into a set of 224$\times$224 HDAD blocks, and then our method outputs the binarized HDAD map $ I'^{bin,HDAD}$.

\section{Experimental result} \label{sec:IV}     
  In the first set of experiments, we demonstrate the accuracy, PSNR, and perceptual effect merits of our method relative to the nine comparative methods \cite{N. Otsu}, \cite{W. Niblack}, \cite{J. Sauvola}, \cite{E. Kavallieratou}, \cite{N. R. Howe}, \cite{Y. H. Chiu}, \cite{SSP}, \cite{J. Calvo}, \cite{J. Zhao}. Here, as an objective quality metric, the definition of PSNR is defined by

  \begin{equation}
    \label{eq:PSNR}
    PSNR = 10\log_{10}\left ( \frac{255^2} {MSE} \right )\\
  \end{equation}

  \noindent with $MSE = \frac{\sum_{x=1}^{W}\sum_{y=1}^{H} (({I}^{bin,HDAD}(x, y)-{I'}^{bin,HDAD}{^(x, y)})^{2}}{WXH}$, in which $I^{bin,HDAD}(x, y)$ and $I'^{bin,HDAD}(x, y)$ denote the ground truth binarized HDAD pixel value, whose value is ``0'' for foreground and ``255'' for background, and the binarized HDAD pixel at the location $(x, y)$ by using the considered method, respectively. “W$\times$H” denotes the size of the HDAD map. In the second set of experiments, we retrain the two state-of-the-art methods \cite{J. Calvo}, \cite{J. Zhao} based on the same HDAD-pair dataset, and then based on the same testing dataset, with the similar binarization accuracy and quality, the experimental data demonstrates the significant execution-time and parameters reduction merits of our method relative to the retrained version of the two methods \cite{J. Calvo}, \cite{J. Zhao}.

  All experiments are implemented using a desktop with an Intel Core i7-7700 CPU running at 3.6 GHz with 24 GB RAM and a Nvidia 1080Ti GPU. The operating system is Microsoft Windows 10 64-bit. The program development environment is Visual Studio 2017 with Python programming language.

  \subsection{Accuracy comparison} \label{sec:IVA}
  To compare the binarization accuracy among the considered methods, the four used metrics are ``Recall'', ``Specificity'', ``Precision'', and ``F-measure''. Because the four metrics involve the four parameters, namely true positive (TP), true negative (TN), false positive (FP), and false negative (FN), they are defined by

  \begin{description}
  \item TP: the number of pixels that are correctly binarized as foreground pixels.
  \item TN: the number of pixels that are correctly binarized as background pixels.
  \item FP: the number of pixels that are erroneously binarized as foreground pixels.
  \item FN: the number of pixels that are erroneously binarized as background pixels.
  \end{description}

  According to the above four basic parameters, the four performance evaluation metrics are defined by
  \begin{equation}
    \label{eq:recall}
    Recall\ (Re) = \frac{TP}{TP+FN}\\
  \end{equation}

  \begin{equation}
    \label{eq:specificity}
    Specificity\ (Sp) = \frac{TN}{TN+FP}\\
  \end{equation}

  \begin{equation}
    \label{eq:precision}
    Precision\ (Pr) = \frac{TP}{TP+FP}\\
  \end{equation}

  \begin{equation}
    \label{eq:F-measure}
    F\mbox{-}measure\ (F\mbox{-}m) = \frac{2}{\frac{1}{recall}+\frac{1}{precision}}\\
  \end{equation}

  Based on the training dataset and the testing dataset \cite{website}, in terms of the four metrics in Eqs. (\ref{eq:recall})-(\ref{eq:F-measure}) and PSNR, in Table \ref{table:subroutines}, we observe that our method has the highest Recall (= 97\%), Specificity (= 99\%), Precision (= 95\%), F-measure (= 96\%), and PSNR (= 24.5737) in boldface, indicating the substantial accuracy superiority of our method over the nine comparative methods in \cite{N. Otsu}, \cite{W. Niblack}, \cite{J. Sauvola}, \cite{N. R. Howe}, \cite{E. Kavallieratou}, \cite{Y. H. Chiu}, \cite{SSP}, \cite{J. Calvo}, \cite{J. Zhao}. For simplicity, the method proposed by Sauvola and Pietikäinen \cite{J. Sauvola} is abbreviated as the SP method; the method proposed by Calvo-Zaragoza and Gallego \cite{J. Calvo} is abbreviated as the CG method. The available code of our CNN-based binarization method can be accessed from the website \cite{website}.

  Since the available codes in the two state-of-the-art methods \cite{J. Calvo}, \cite{J. Zhao} are provided for the DIBCO dataset \cite{DIBCO}, for completeness and fairness, we retrain the two methods. In Table \ref{table:CNN_compar}, we observe that our method has competitive accuracy and PSNR relative to the two considered methods, but using our method, the number of required parameters and the execution-time requirement (in seconds) can be significantly reduced, thus providing a better opportunity to embed our method into embedding systems.\\

  \begin{table*}[] 
  \caption{Accuracy and PSNR merits of our method relative to the previous nine methods with available codes.}
  \resizebox{\textwidth}{!}{
  \begin{tabular*}{{11cm}}{ c c c c c c } 
  \toprule
  Method         & Re          & Sp          & Pr          & F-m & PSNR        \\ \hline
  \midrule
      Otsu \cite{N. Otsu}                         & 0.7683          & 0.9550          & 0.8173          & 0.7549          & 16.7964         \\ \hline
      Niblack \cite{W. Niblack}                   & 0.7530          & 0.6921          & 0.1451          & 0.2338          & 6.1884         \\ \hline
      SP \cite{J. Sauvola}   & 0.8740          & 0.9528          & 0.6449          & 0.7066          & 15.3962         \\ \hline
      Kavallierataou \cite{E. Kavallieratou}      & 0.9073          & 0.9468          & 0.6496          & 0.7159          & 15.2781         \\ \hline
      Howe \cite{N. R. Howe}                      & 0.7371          & 0.9559          & 0.5963          & 0.6376          & 14.2637         \\ \hline
      Chiu $et$ $al$. \cite{Y. H. Chiu}           & 0.8948          &0.9929  & 0.9154          & 0.9033          & 20.9130         \\ \hline
      Jia $et$ $al$. \cite{SSP}                   & 0.8570          &0.9876  & 0.8889          & 0.8717          & 19.6978         \\ \hline
      CG \cite{J. Calvo}  & 0.7380          &0.9873  & 0.8317          & 0.7047          & 17.5464         \\ \hline
      Zhao $et$ $al$. \cite{J. Zhao}              & 0.6102          & 0.9940          & 0.9043          & 0.6973          & 16.9838         \\ \hline
      Proposed method                             & \textbf{0.9680} & \textbf{0.9950} & \textbf{0.9522} & \textbf{0.9597} & \textbf{24.5737} \\ \hline
  \bottomrule
  \label{table:subroutines}
  \end{tabular*}}
  \end{table*}

  \begin{table*}[]  
  \caption{With competitive quality and PSNR, the parameters and execution-time reduction merits of our method relative to the retrained version of two state-of-the-art CNN-based methods \cite{J. Calvo}, \cite{J. Zhao}.}
  \resizebox{\textwidth}{!}{
  \begin{tabular*}{17cm}{ c c c c c c c c }  
  \toprule
    Method & Re & Sp & Pr & F-m & PSNR  &  \#(parameters) & time (s) \\ \hline
  \midrule
    CG \cite{J. Calvo} & 0.9635   & \textbf{0.9970}   & \textbf{0.9646}   & \textbf{0.9638} & 23.4270    & 928,001   & 21.68          \\ \hline
    Zhao $et$ $al$. \cite{J. Zhao}           & 0.9605   & 0.9966   & \textbf{0.9633}   & 0.9618 & \textbf{24.6297}     & 26,790,788  & 22.53         \\ \hline
    Proposed method                   & \textbf{0.9680}   & 0.9950   & 0.9522   & 0.9597 &24.5737     & \textbf{84,654}  & \textbf{4.19}              \\ \hline
  \bottomrule
  \label{table:CNN_compar}
  \end{tabular*}}
  \end{table*}

  \begin{figure*}[!h]
      \centering
          \subfigure[]{\includegraphics[width=0.13\linewidth]{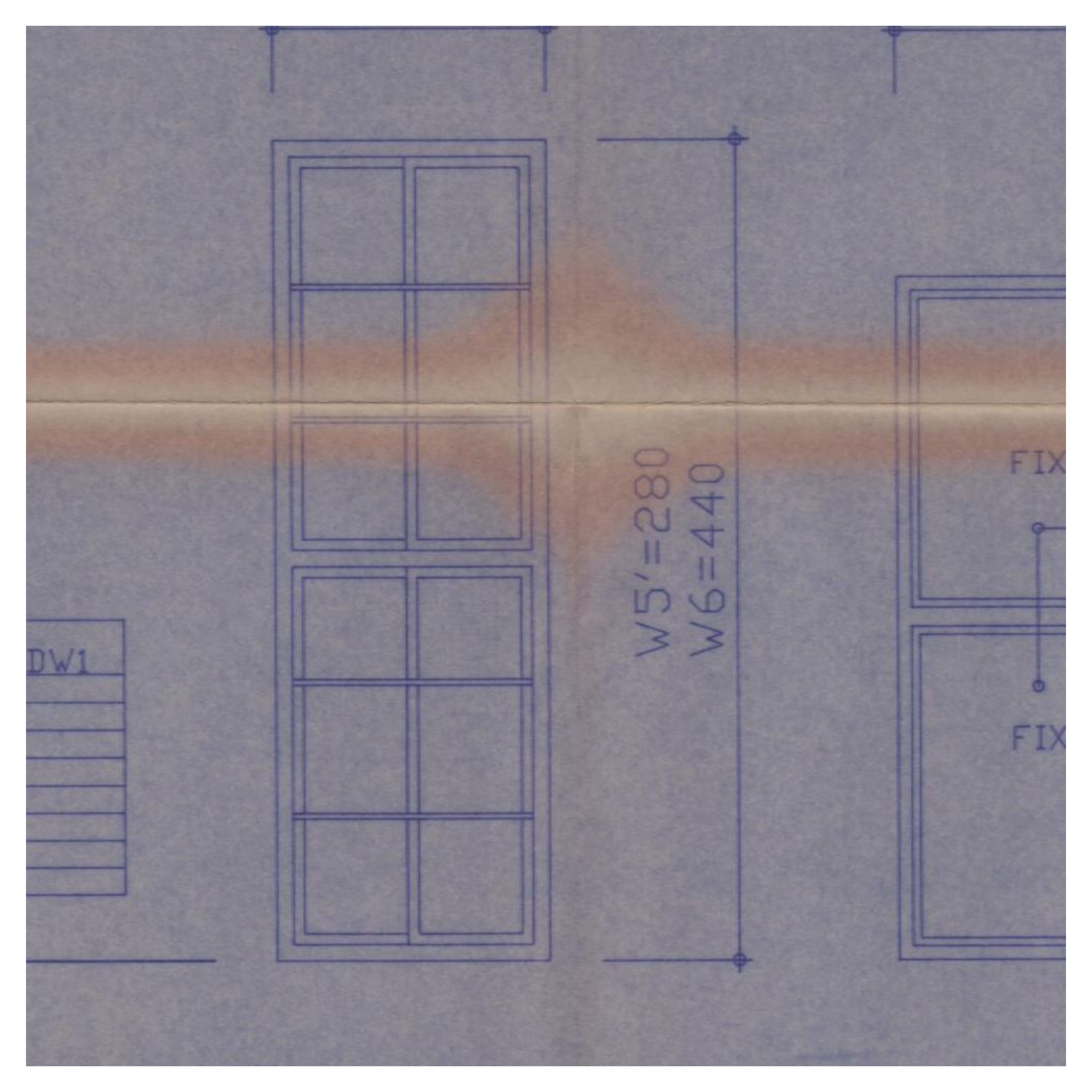}}
      \centering
          \subfigure[]{\includegraphics[width=0.13\linewidth]{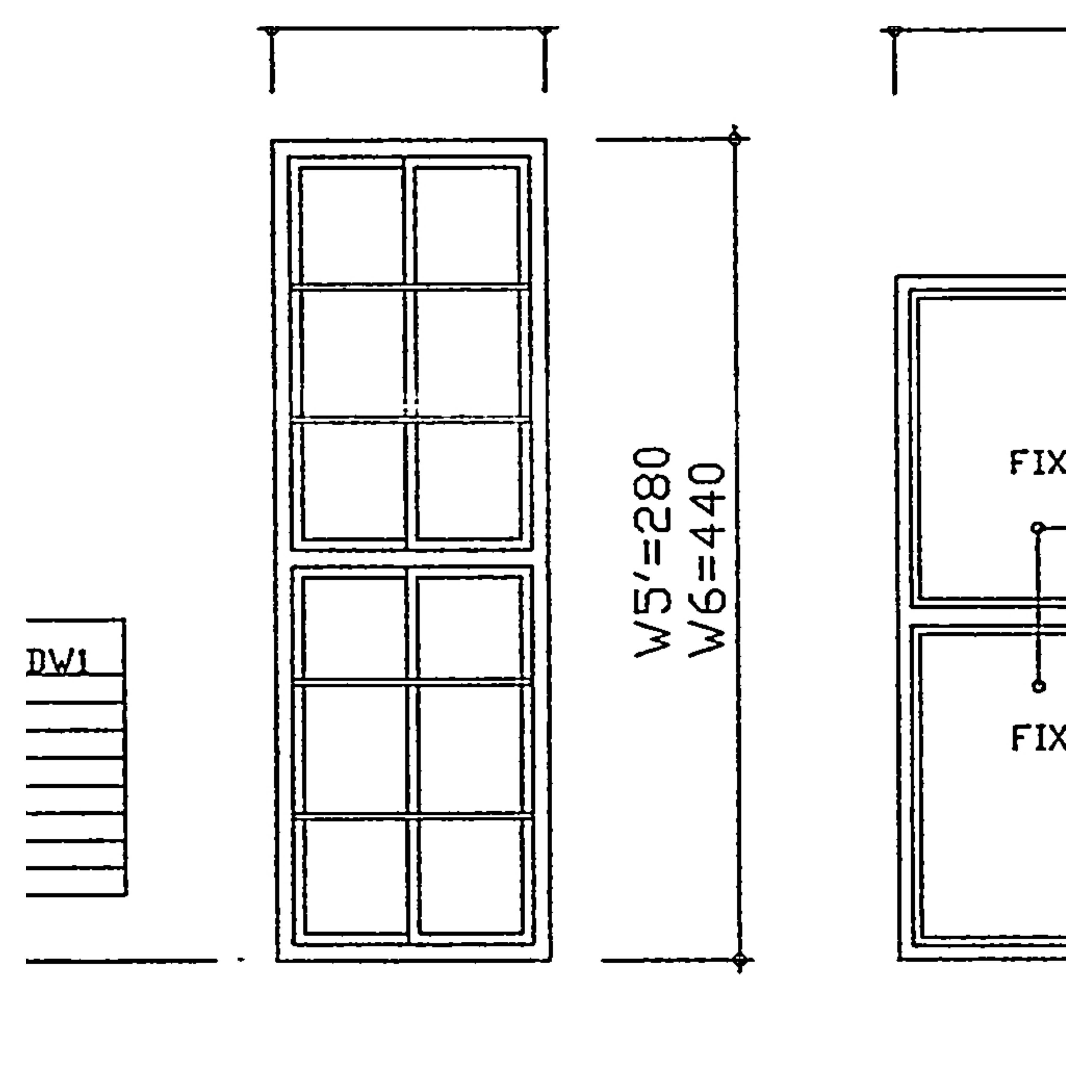}}
      \centering
          \subfigure[]{\includegraphics[width=0.13\linewidth]{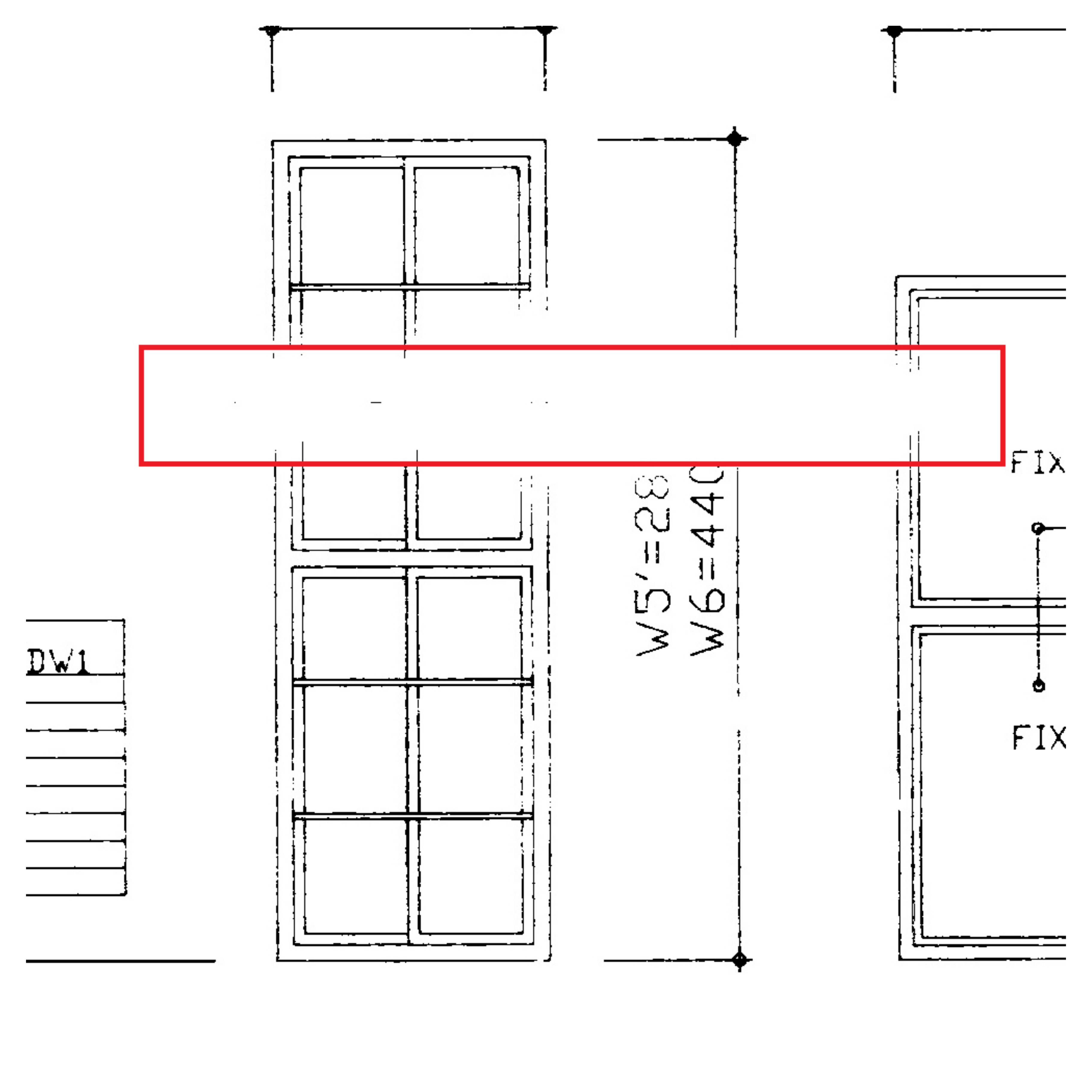}}
      \centering
          \subfigure[]{\includegraphics[width=0.13\linewidth]{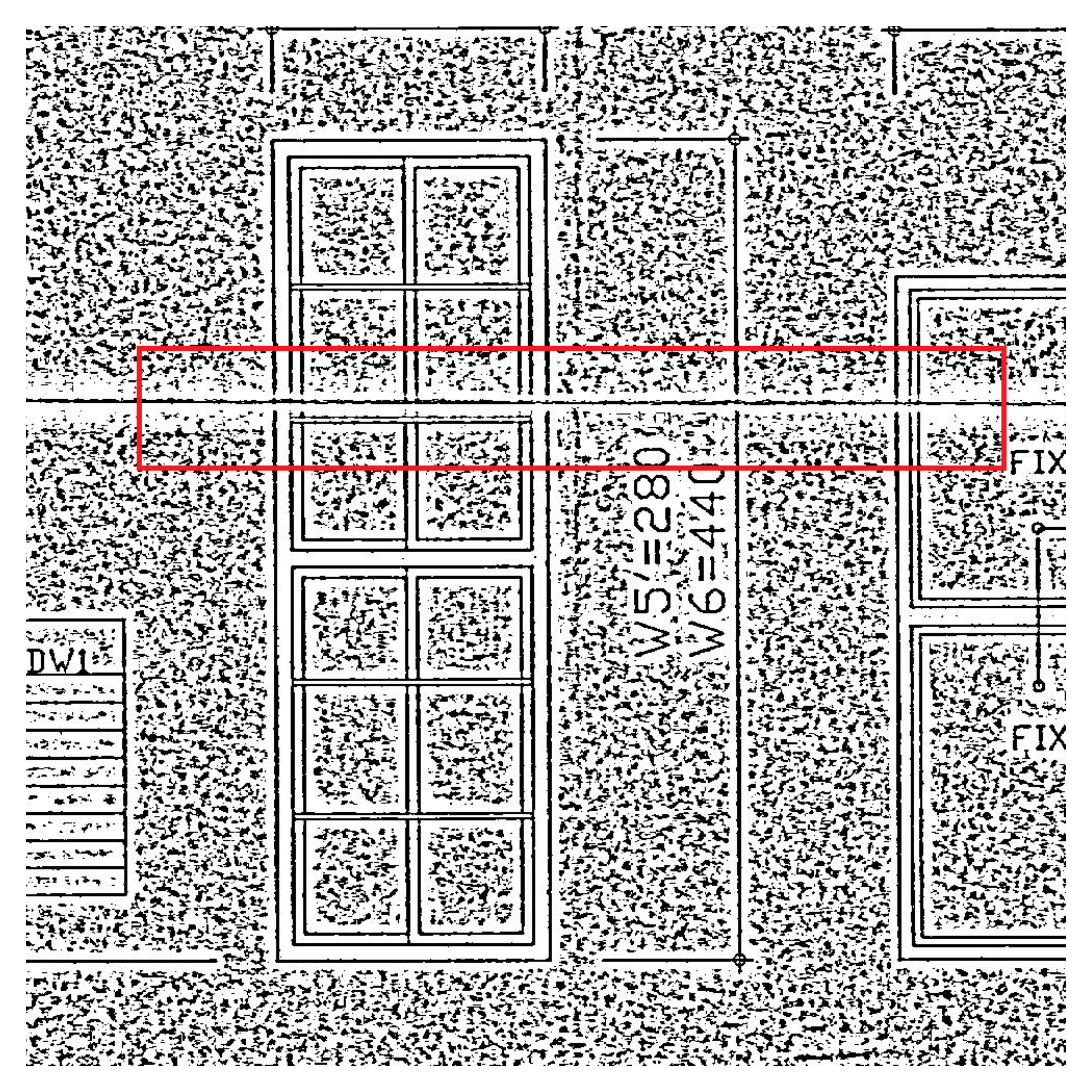}}
      \centering
          \subfigure[]{\includegraphics[width=0.13\linewidth]{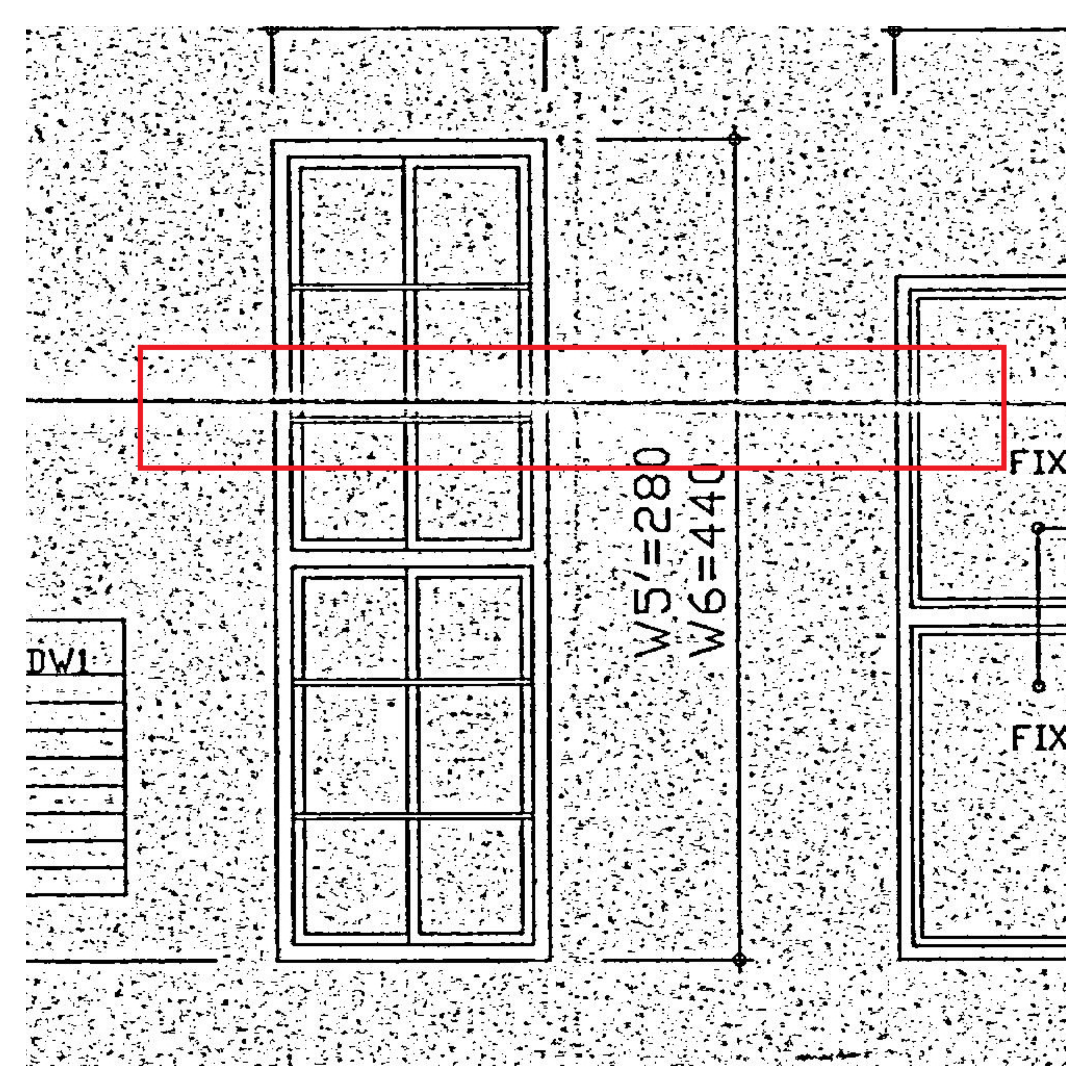}}
      \centering
          \subfigure[]{\includegraphics[width=0.13\linewidth]{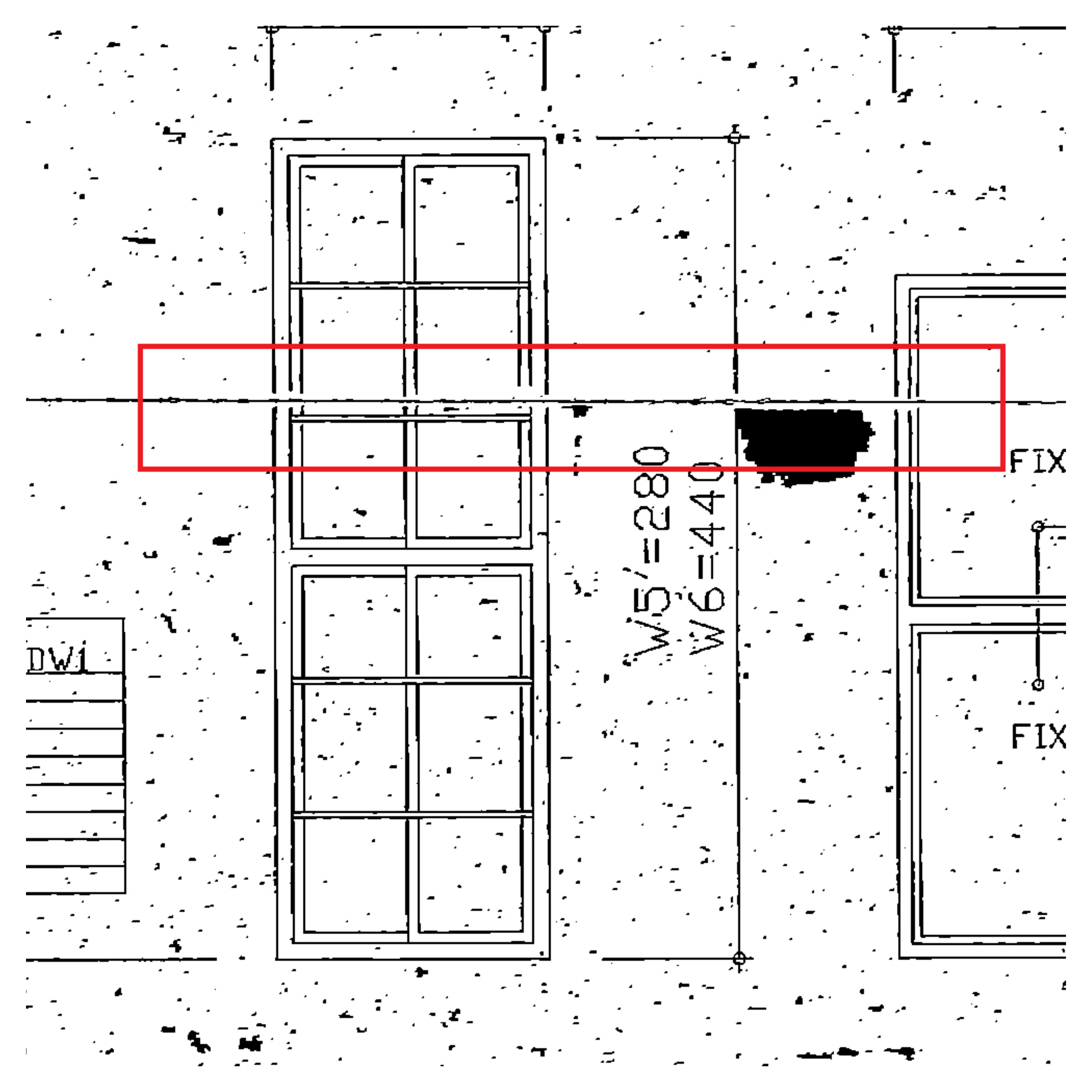}}
      \centering
          \subfigure[]{\includegraphics[width=0.13\linewidth]{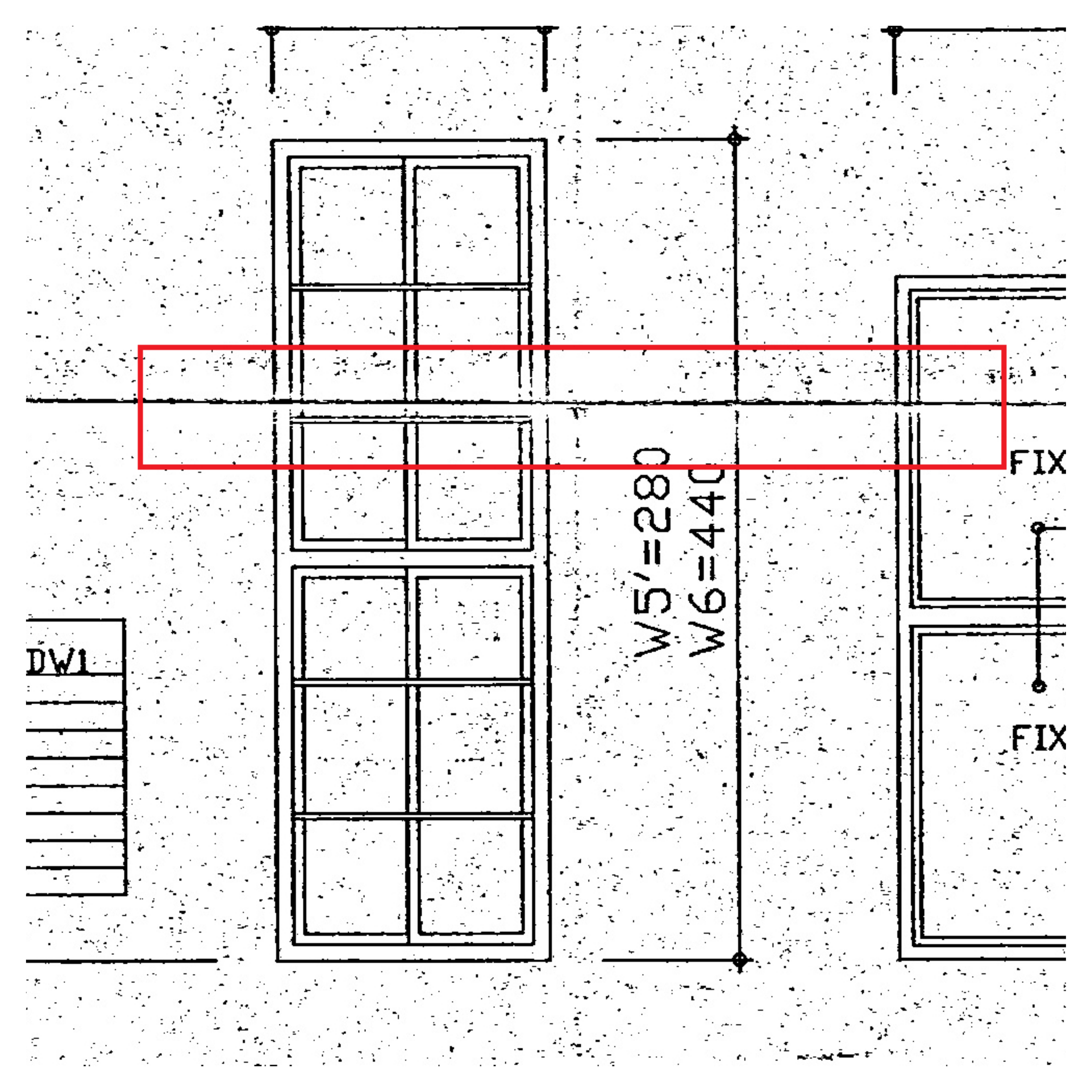}}
      \centering
          \subfigure[]{\includegraphics[width=0.135\linewidth]{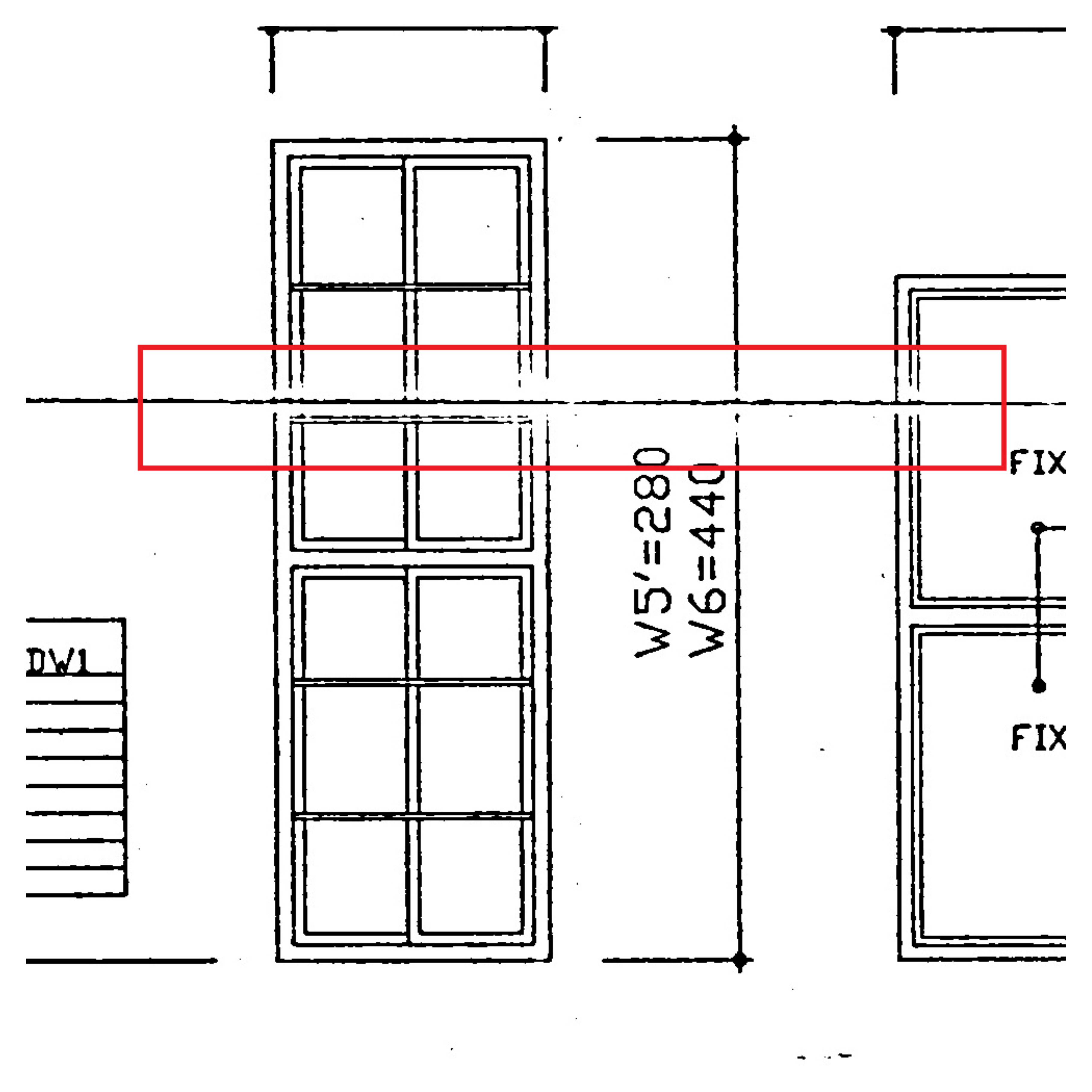}}
      \centering
          \subfigure[]{\includegraphics[width=0.135\linewidth]{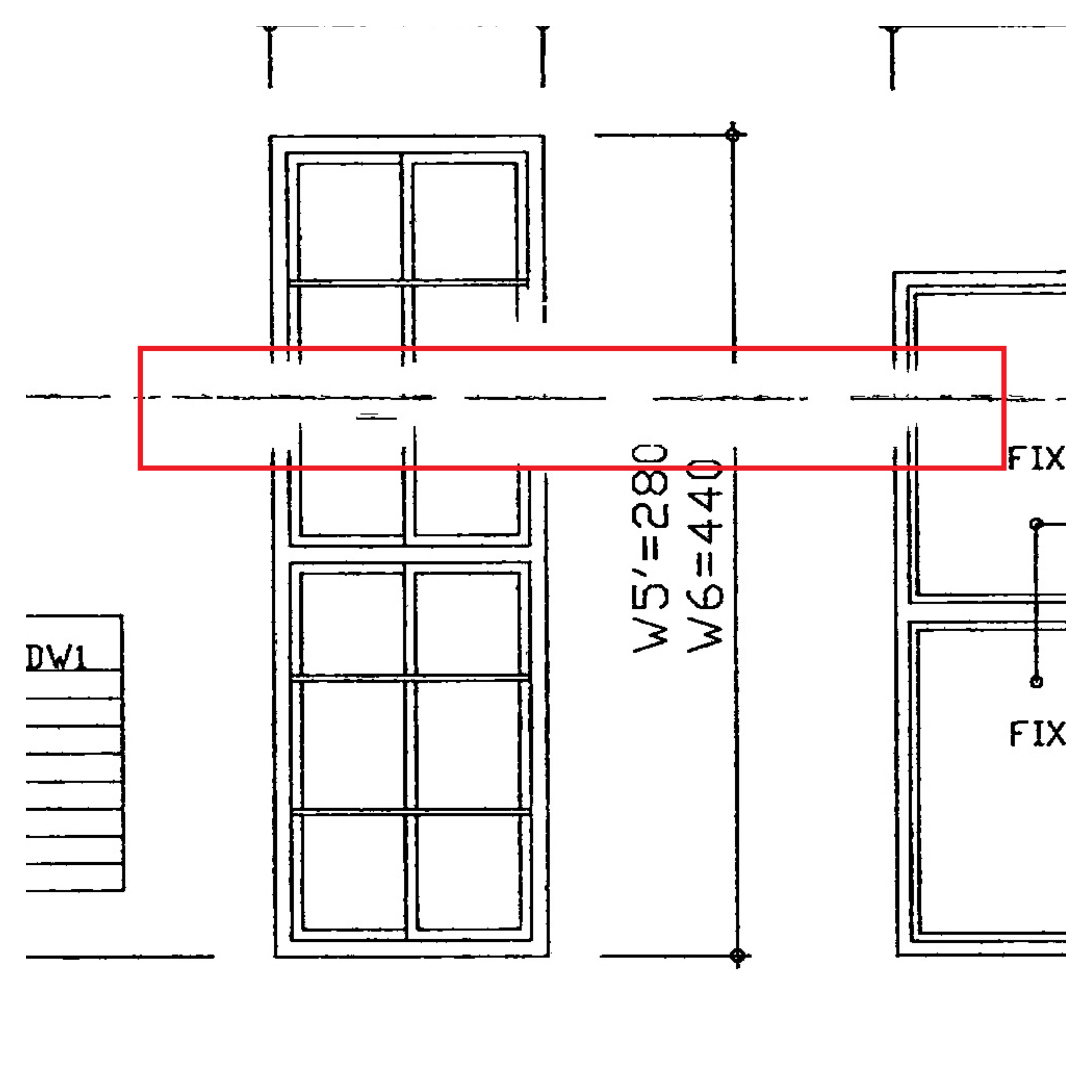}}
      \centering
          \subfigure[]{\includegraphics[width=0.135\linewidth]{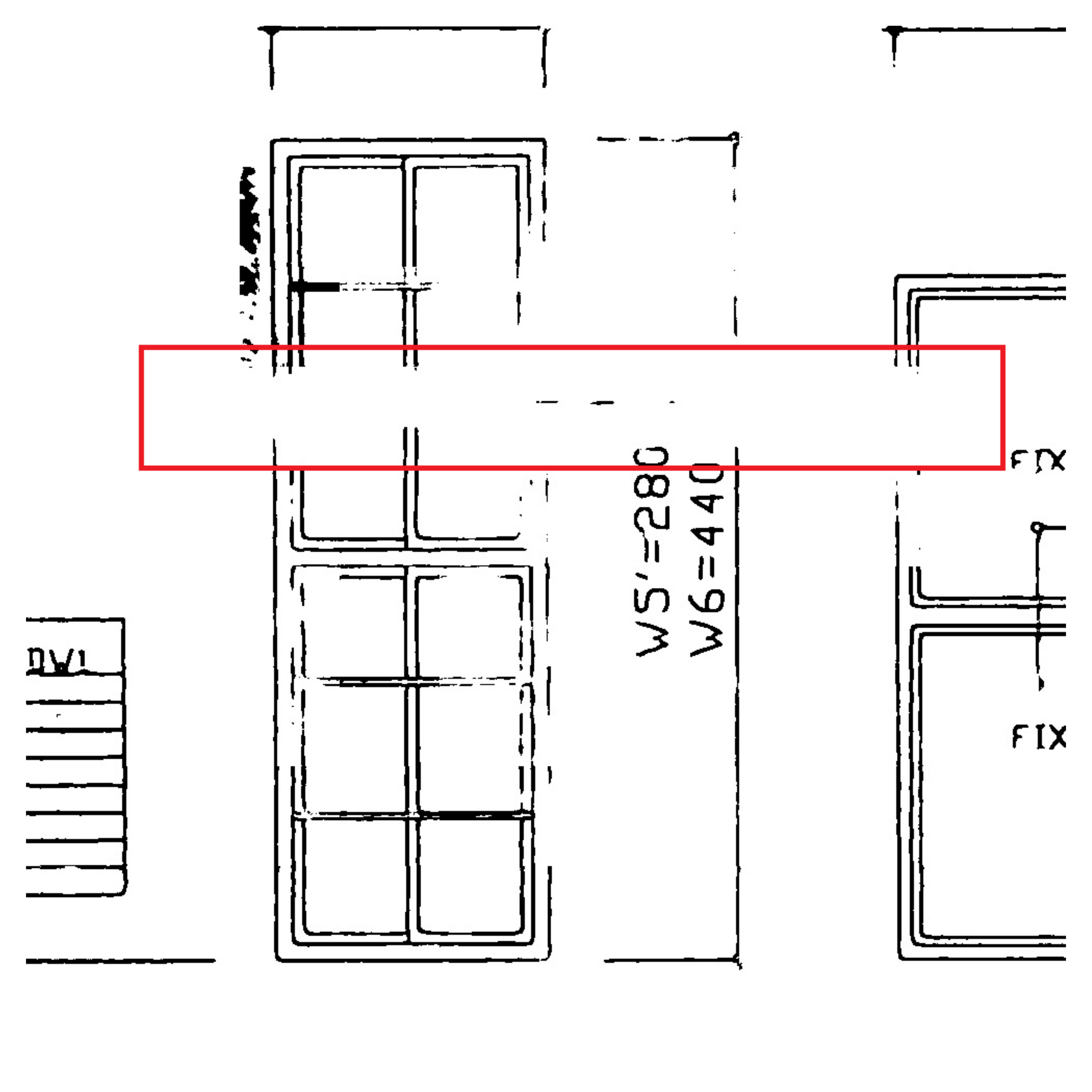}}
      \centering
          \subfigure[]{\includegraphics[width=0.135\linewidth]{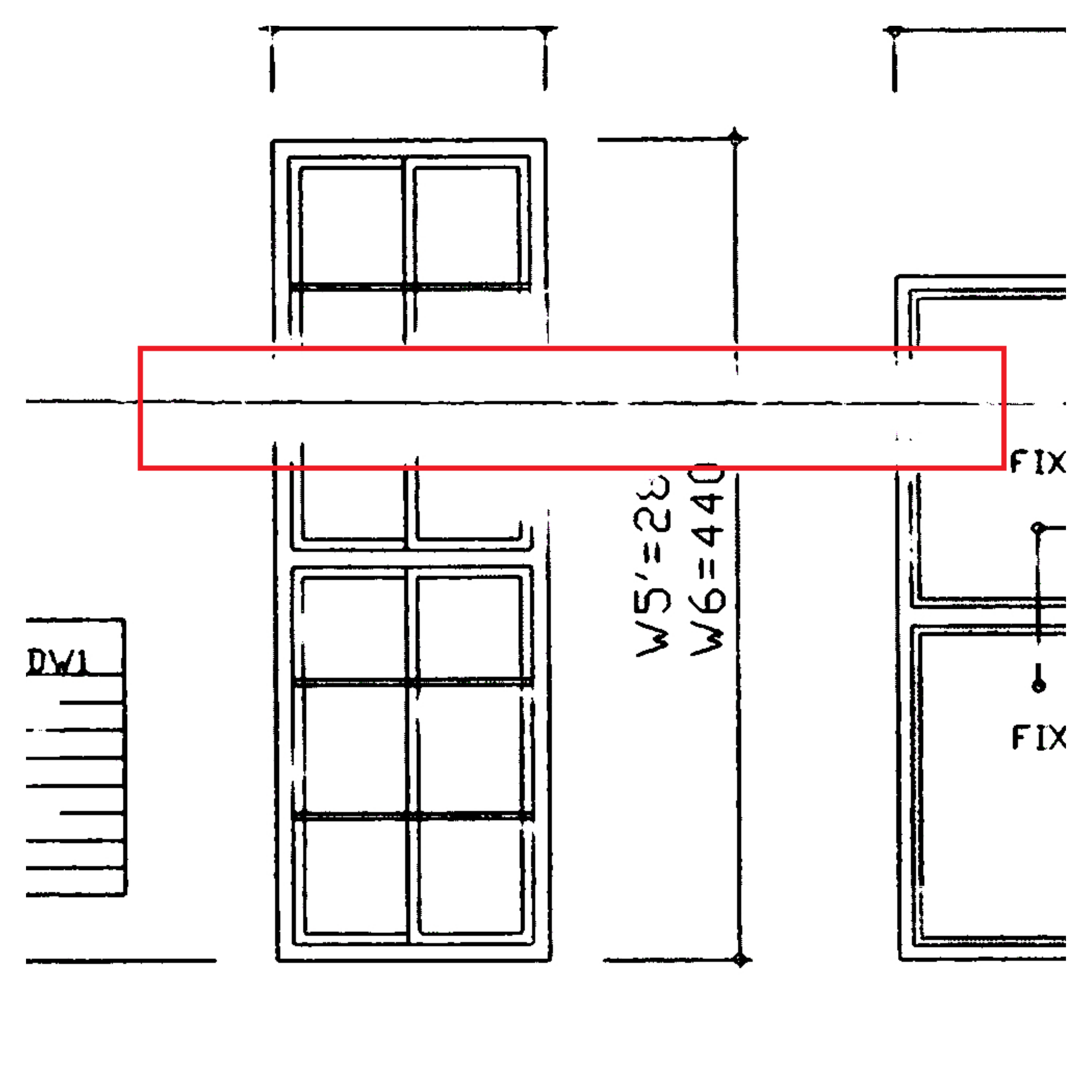}}
      \centering
          \subfigure[]{\includegraphics[width=0.135\linewidth]{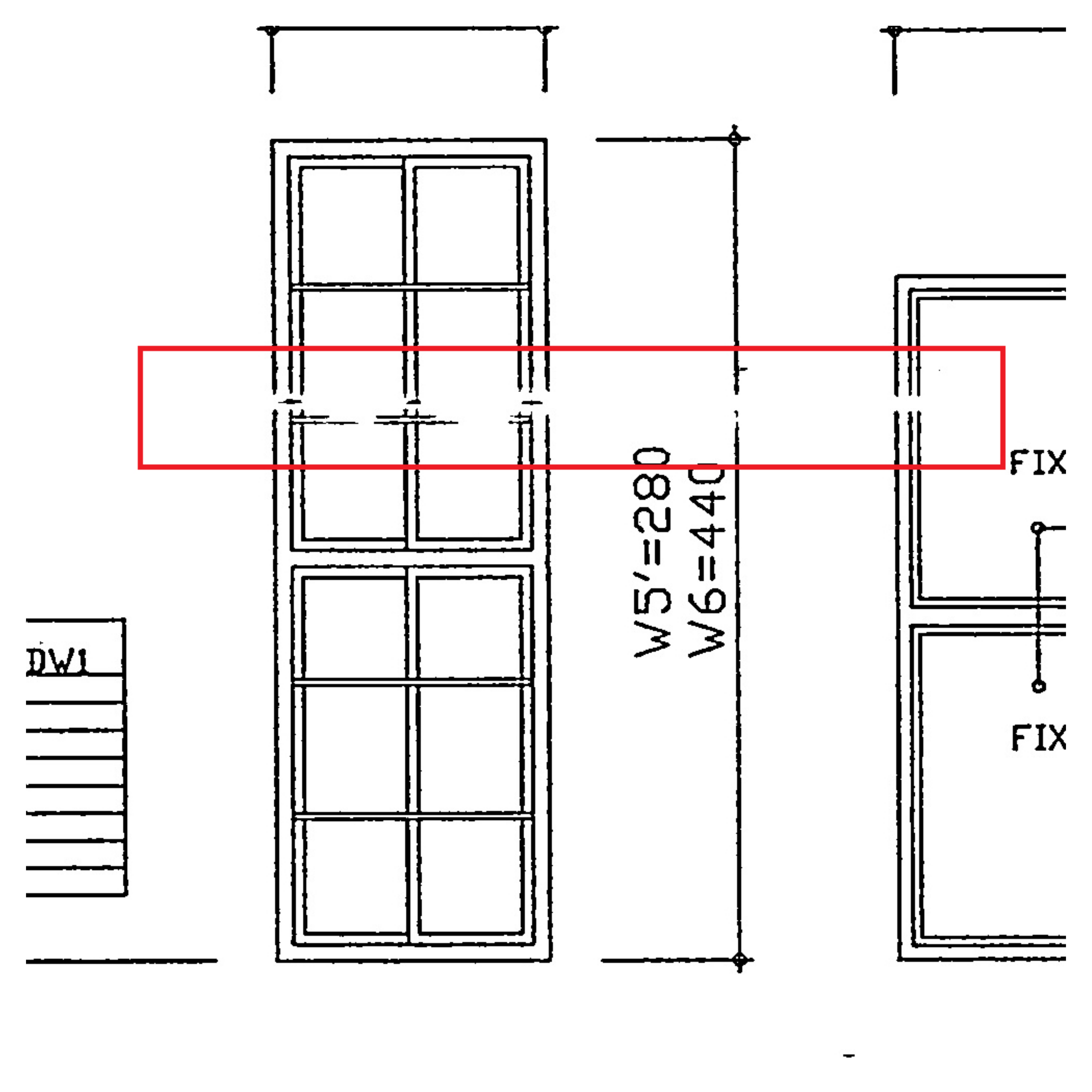}}
      \centering
          \subfigure[]{\includegraphics[width=0.135\linewidth]{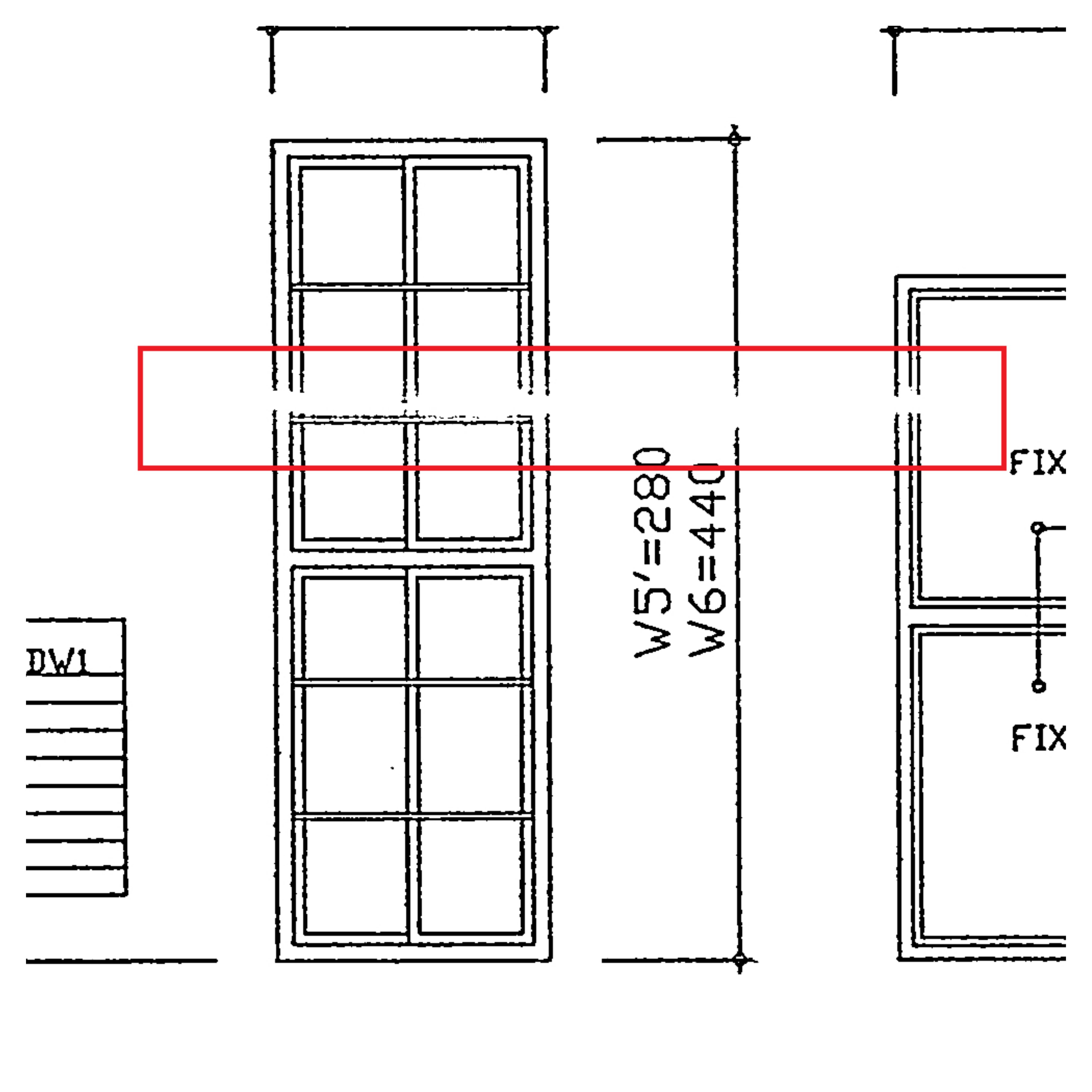}}
      \centering
          \subfigure[]{\includegraphics[width=0.135\linewidth]{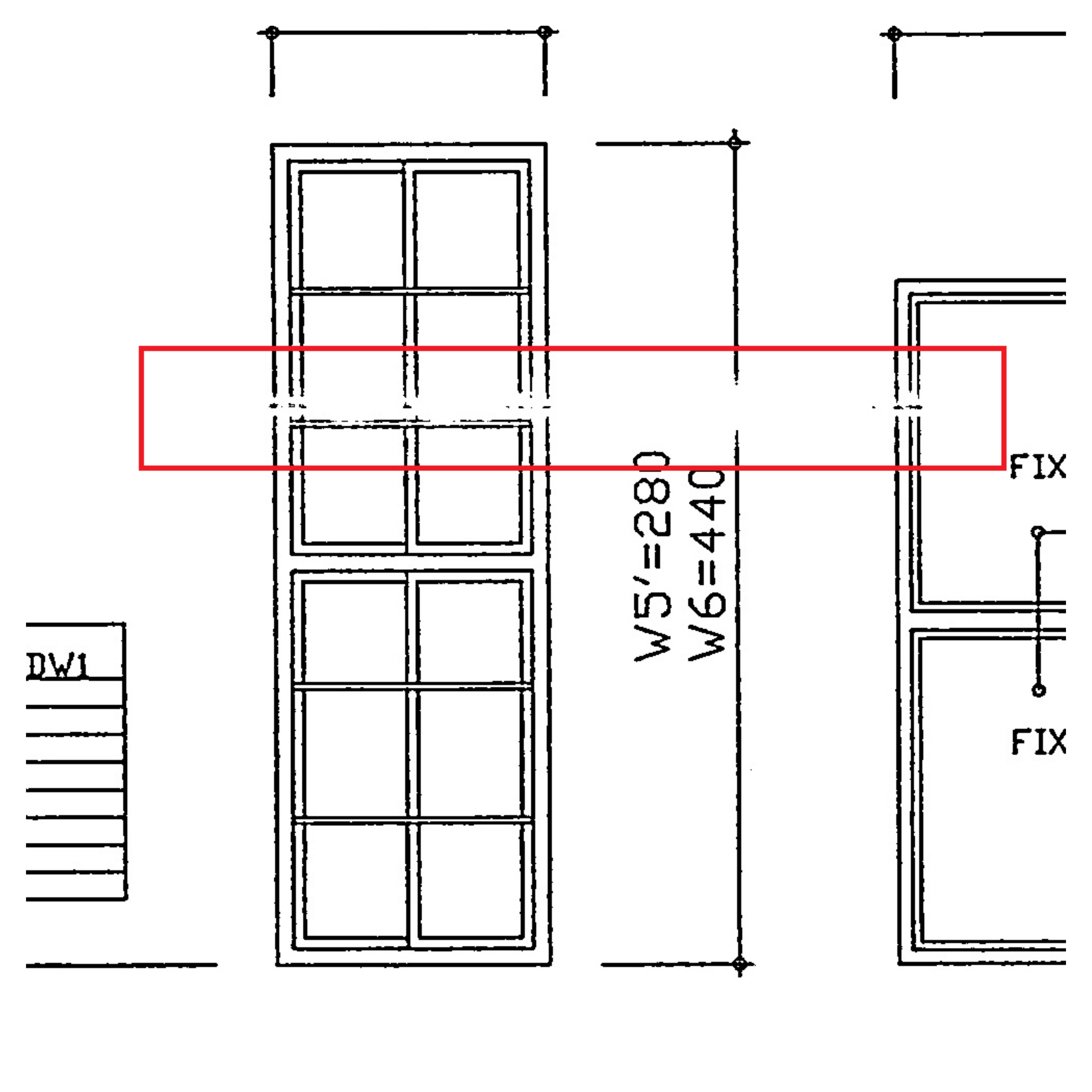}}
      \caption{The perceptual effect merit of the proposed method. (a) Original HDAD map. (b) The ground truth binarized map of Fig. \ref{fig:1068}(a). (c) Otsu \cite{N. Otsu}. (d) Niblack \cite{W. Niblack}. (e) Sauvola and Pietikäinen \cite{J. Sauvola}. (f) Howe \cite{N. R. Howe}. (g) Kavallieratou \cite{E. Kavallieratou}. (h) Chiu $et$ $al$. \cite{Y. H. Chiu}. (i) Jia $et$ $al$. \cite{SSP}. (j) Calvo-Zaragoza and Gallego \cite{J. Calvo}. (k) Zhao $et$ $al$. \cite{J. Zhao}. (l) Retrained version of Calvo-Zaragoza and Gallego's method \cite{J. Calvo}. (m) Retrained version of Zhao $et$ $al$.'s method \cite{J. Zhao}. (n) Proposed method.}
      \label{fig:1068}
  \end{figure*}

  \subsection{Perceptual effect merit}
  To show the subjective quality merit of our method by human eyes, we take the HDAD map in Fig. \ref{fig:1068}(a) as the example to demonstrate the perceptual effect merit of our method. As the comparison base, Fig. \ref{fig:1068}(b) illustrates the ground truth binarized result of Fig. \ref{fig:1068}(a). After performing the nine comparative methods \cite{N. Otsu}, \cite{W. Niblack}, \cite{J. Sauvola}, \cite{N. R. Howe}, \cite{E. Kavallieratou}, \cite{Y. H. Chiu}, \cite{SSP}, \cite{J. Calvo}, \cite{J. Zhao}, the retrained version of the two state-of-the-art methods \cite{J. Calvo}, \cite{J. Zhao}, and our method on Fig. \ref{fig:1068}(a), Figs. \ref{fig:1068}(c)-(n) depict the twelve binarized HDAD maps. Among these binarized HDAD maps, we observe that our method has better perceptual effect than that by using the nine comparative methods. When compared with the retrained version of the two methods \cite{J. Calvo}, \cite{J. Zhao}, our method has similar perceptual effect; however, as demonstrated in Table \ref{table:CNN_compar}, our method has much less parameters and execution-time requirements, providing a better opportunity to embed our method into embedding systems.

\section{Conclusion} \label{sec:V}
Prior to manipulating digital HDAD maps, binarizing HDAD maps is a necessary step. Due to noise, the yellowing area artifact, the folded line artifact, and the complicated foreground components, the binarization of HDAD maps is a new challenging job. We have presented our labeling and CNN-based binarization method for HDAD maps. First, we propose a semi-automatic labeling method to produce the ground truth HDAD-pair dataset effectively. Secondly, based on the newly created HDAD-pair dataset, we propose an effective CNN-based binarization method for HDAD maps. The thorough experimental data have demonstrated the accuracy, PSNR, and the perceptual effect merits of our method relative to the nine comparative methods \cite{N. Otsu}, \cite{W. Niblack}, \cite{J. Sauvola}, \cite{N. R. Howe}, \cite{E. Kavallieratou}, \cite{Y. H. Chiu}, \cite{SSP}, \cite{J. Calvo}, \cite{J. Zhao}. Note that in terms of the same quality metrics, the proposed method in this paper is much superior to our preliminary \cite{K. L. Chung}. Although our method has competitive quality, PSNR, and the perceptual effect as that of the retrained version of the methods \cite{J. Calvo}, \cite{J. Zhao}, our method has significant execution-time and parameters reduction merits, providing a better an opportunity to embed into embedding systems.

\section{Acknowledgement} \label{sec:VI}
The authors appreciate the proofreading help of Ms. Catherine Harrington to improve the manuscript. This work was supported by Grants MOST-107-2221-E-011-108-MY3 and MOST-108-
2221-E-011-077-MY3.\\

{
\footnotesize
\bibliographystyle{ieee}
\bibliography{YL_arxiv_release}
}

\end{document}